\documentclass[10pt,journal,letterpaper,compsoc]{IEEEtran}
    \newtheorem{theorem}{Theorem}[section]
    
    \newtheorem{definition}[theorem]{Definition}
    \newtheorem{proposition}[theorem]{Proposition}

    \newcommand{\qed}{\nobreak \ifvmode \relax \else
          \ifdim\lastskip<1.5em \hskip-\lastskip
          \hskip1.5em plus0em minus0.5em \fi \nobreak
          \vrule height0.75em width0.5em depth0.25em\fi}

\ifCLASSOPTIONcompsoc
\else
\fi

\usepackage{hyperref}
\usepackage{graphicx}
\usepackage{psfrag}
\usepackage{float}

\usepackage{algorithm}
\usepackage{algorithmic}

\usepackage[cmex10]{amsmath}
\usepackage{amssymb}

\usepackage{algorithmic}
\usepackage{enumerate}

\begin{document}
%
\title{On Recursive Edit Distance Kernels with Application to Time Series Classification}

\author{Pierre-Fran\c{c}ois~Marteau,~\IEEEmembership{Member,~IEEE} and Sylvie~Gibet,~\IEEEmembership{Member,~IEEE}

\IEEEcompsocitemizethanks{\IEEEcompsocthanksitem P.F. Marteau and Sylvie Gibet are with the  Univ. Bretagne Sud, UMR 6074, IRISA, F-56000 Vannes, France.\protect\\
E-mail: \{Pierre-Francois.Marteau, Sylvie.Gibet\}(AT)univ-ubs.fr 
}
\thanks{}}

\markboth{Draft paper in submission}%
{Marteau P.F. and Gibet S. \MakeLowercase{\textit{}}: On Recursive Edit Distance Kernels with Application to Time Series Classification}

\IEEEcompsoctitleabstractindextext{%
\begin{abstract}

This paper proposes some extensions to the work on kernels dedicated to string or time series global alignment based on the aggregation of scores obtained by local alignments. The extensions that we propose allow to construct, from classical recursive definition of elastic distances, recursive edit distance (or time-warp) kernels that are positive definite if some sufficient conditions are satisfied. The sufficient conditions we end-up with are original and weaker than those proposed in earlier works, although a recursive regularizing term is required to get the proof of the positive definiteness as a direct consequence of the Haussler's convolution theorem. Furthermore, the positive definiteness is maintained when a symmetric corridor is used to reduce the search space and thus the algorithmic complexity, which, is quadratic in the worse case . The classification experiment we conducted on three classical time warp distances (two of which being metrics), using Support Vector Machine classifier, leads to the conclusion that, when the pairwise distance matrix obtained from the training data is \textit{far} from definiteness, the positive definite recursive elastic kernels outperform in general the distance substituting kernels for several classical elastic distances we have tested. 
\end{abstract}

\begin{IEEEkeywords}
Edit distance, Dynamic Time Warping, Recursive kernel, Time series classification, Support Vector Machine, Definiteness.
\end{IEEEkeywords}}

\maketitle

\IEEEdisplaynotcompsoctitleabstractindextext

\IEEEpeerreviewmaketitle

\section{Introduction}
\IEEEPARstart{E}{lastic} similarity measures such as Dynamic Time Warping (DTW) or Edit Distances have proved  to be quite efficient compared to non-elastic similarity measures such as Euclidean measures or LP norms when addressing tasks that require the matching of time series data, in particular time series clustering and classification. A  wide scope of applications in different domains, such as physics, chemistry, finance, bio-informatics, network monitoring, etc, have demonstrated the benefits of using elastic measures. 
A natural question to ask at this point is whether we can develop and use \textit{elastic} kernel methods based from such elastic distances. This requires to address the existence and construction of Reproducing Edit Distance Hilbert Spaces for a given elastic measure, i.e. a functional vector space endowed with the dot product in which edit distances or time warping algorithms can be defined. Unfortunately it seems that common elastic measures that are derived from DTW or more generally dynamic programming recursive algorithms are not directly induced by an inner product of any sort, even when such measures are metrics. One can conjecture that it is not possible to embed time series in an Hilbert space having a time-warp capability using these classical elastic measures, but nevertheless, we can propose (close) regularized variants for which such kernel construction construction is possible.

This paper aims at (re-)exploring this issue and, following earlier works (\cite{TWED:Haussler99}, \cite{TWED:Haasdonk04}, \cite{TWED:Vert2004}, \cite{TWED:Cortes04}, \cite{TWED:Cuturi07})  proposes Recursive Edit Distance Kernels (REDK) constructions that try to preserve the properties of elastic measures from which they are derived, while offering the possibility of embedding time series in Time Warped Hilbert Spaces. 
The main contributions of the paper are as follows
\begin{enumerate}
\item we verify the indefiniteness of the main time-warp measures used in the literature,
\item we propose a new method to construct positive definite kernels from classical time-warp or edit measures. This method is quite general and less restrictive than previous ones, although it requires to introduce a recursive regularizing term that allows to prove the definiteness of our kernels as a direct consequence of the Haussler's convolution theorem, 
\item we show that the regularizing approach we develop applies also when a symmetric \textit{corridor} is used to limit the search space used to evaluate the elastic measure,
\item we experiment and compare the proposed kernels on some time series classification tasks using a large variety of time series data sets to estimate in practice the benefit we can expect from such kernels.\\
\end{enumerate}

The paper is organized as follows: the second section synthesizes the related works; the third section introduces the notation and mathematical background that is used throughout the paper. The fourth section develops the construction of a general REDK and details some instantiations derived from classical elastic measures. The fifth section gathers classification experimentation on a wide range of time series data and compares REDK accuracies with classical elastic measures. The sixth section proposes a conclusion and  further research perspectives. Appendix \ref{Appendix:A} states the indefiniteness of classical elastic measures, and appendix \ref{Appendix:B} presents the proof of our main results.

\section{Related works}

During the last decades, the use of kernel-based methods in pattern analysis has provided numerous results and fruitful applications in various domains such as biology, statistics, networking, signal processing, etc. Some of these domains, such as bioinformatics, or more generally domains that rely on sequence or time series models, require the analysis and processing of variable length vectors, sequences or timestamped data.
Various methods and algorithms have been developed to quantify the similarity of such objects. From the original dynamic programming \cite{TWED:Bellman57} implementation of the  symbolic edit distance \cite{TWED:Levenshtein66} by Wagner and Fisher \cite{TWED:Wagner73}, the Smith and Waterman (SW) algorithm \cite{TWED:SmithWaterman81} has been designed to evaluate the similarity between two symbolic sequences by means of a local gap alignment. More efficient local heuristics have since been proposed to meet the massive symbolic data challenge, such as BLAST \cite{TWED:Altschul90} or FASTA \cite{TWED:Pearson90}. Similarly, dynamic time warping measures have been developed to evaluate similarity between digital time series or timestamped data \cite{TWED:Velichko70}, \cite{TWED:Sakoe71}, and more recently \cite{TWED:Chen04}, \cite{TWED:Marteau09} propose elastic metrics dedicated to such digital data.

Our ability to construct kernels with elastic or time-warp properties from such \textit{elastic distances} allowing to embed time series into vector spaces (Euclidean or not) has attracted attention (e.g. \cite{TWED:Haasdonk04}\cite{TWED:Hayashi2005}\cite{TWED:Haasdonk05}). Indeed, significant benefits can be expected from applications of kernel-based machine learning algorithms to variable length data, or more generally data for which some elastic matching has a meaning. Among the kernel machine algorithms applicable to discrimination or regression tasks, Support Vector Machines (SVM) \cite{TWED:Vapnik89}, \cite{TWED:BoserGV92}, \cite{TWED:Scholkopf01} are still reported to yield state-of-the art performances although their accuracy greatly depends on the exploited kernel.

The definition of \textit{good} kernels from known elastic or time-warp distances applicable to data of variable lengths has been a major challenge since the 1990s. The notion of 'goodness' has rapidly been associated to the concept of definiteness. Basically SVM algorithms involve an optimization process whose solution is proved to be uniquely defined if and only if the kernel is positive definite: in that case the objective function to optimize is quadratic and the optimization problem convex. Nevertheless, if the definiteness of kernels is an issue, in practice, many situations exist where definite kernels are not applicable. This seems to be the case for the main elastic measures traditionally used to estimate the similarity of objects of variable lengths. A pragmatic approach consists of using indefinite kernels \cite{TWED:Haasdonk04}\cite{TWED:Haasdonk05}\cite{Zhang:2010}, although contradictory results have been reported about the impact of definiteness or indefiniteness of kernels on the empirical performances of SVMs. The sub-optimality of the non-convex optimization process is possibly one of the causes leading to these un-guaranteed performances \cite{TWED:Woznica2004}, \cite{TWED:Haasdonk04}.  Regulation procedures have been proposed to locally approximate indefinite kernel functions by definite ones with some benefits. Among others, some approaches apply direct spectral transformations to indefinite kernels. These methods \cite{TWED:Wu2005} \cite{Chen:jmlr2009} consist in i) flipping the negative eigenvalues or shifting the eigenvalues using the minimal shift value required to make the spectrum of eigenvalues positive, and ii) reconstructing the kernel with the original eigenvectors in order to produce a positive semidefinite kernel. Yet, in general, 'convexification' procedures are difficult to interpret geometrically and the expected effect of the original indefinite kernel may be lost.  Some theoretical highlights have been provided through approaches that consist in embedding the data into a pseudo-Euclidean (pE) space and in formulating the classification problem with an indefinite kernel, such as  minimizing the distance between convex hulls formed from the two categories of data embedded in the pE space \cite{TWED:Haasdonk05}. The geometric interpretation results in a constructive method allowing for the understanding, and in some cases the prediction of the classification behavior of an indefinite kernel SVM in the corresponding pE space. 

Some other works like \cite{TWED:Sivaramakrishnan2004}, \cite{TWED:Kumara2008} address the construction of elastic kernels for time series analysis through a decomposition of time series as a sum of local low degree polynomials and, using a resampling process, the piece-wise approximation of the time series are embedded into a proper so-called Reproducing Kernel Hilbert Space in which the SVM is learned.

Other approaches try to use directly the elastic distance into the kernel construction, without any approximation or resampling process.These works are based on the work of Haussler on convolution kernels \cite{TWED:Haussler99} defined on a set of discrete structures such as strings, trees, or graphs. The iterative method that is developed is generative, as it allows for the building of complex kernels from the convolution of simple local kernels. Following the work of Haussler \cite{TWED:Haussler99}, Saigo et al \cite{TWED:Saigo04} define, from the Smith and Waterman algorithm \cite{TWED:SmithWaterman81}, a kernel to detect local alignment between strings by convolving simpler kernels. These authors show that the Smith and Waterman distance measure, dedicated to  determining similar regions between two nucleotide or protein sequences, is not definite, but is nevertheless connected to the logarithm of a point-wise limit of a series of definite convolution kernels. Cuturi et al. \cite{TWED:Cuturi07} have adapted this approach to times series alignments covering the DTW elastic distance. In fact, these previous studies have very general implications, the first one being that classical elastic measures can also be understood as the limit of a series of definite convolution kernels. 


In this paper we tackle a new approach to the positive definiteness regularization of elastic kernels that follows directly the lines of the Haussler's theorem on convolution kernels. The condition to obtain positive definiteness is weaker than the one proposed in previous works \cite{TWED:Cuturi07}, although it requires the addition of an explicit recursive regularizing term. This term is easy to evaluate and has the same complexity than the recursive equation that defines the elastic distance. Our regularization strategy is thus quite general, independent from the data (although a parameter can be optimized), and can be applied to a very large family of editing or dynamic time warping like distances. It applies also to some of their variants which exploit a symmetric \textit{corridor} to speed-up the computation.

\section{Notations and mathematical backgrounds}
We give in this section commonly used definitions, with few details, for metric, kernel and definiteness, sequence set, and classical elastic measures.

\subsection{Kernel and definiteness}
A very large literature exists on kernels, among which \cite{TWED:BergChristensenRessel84}, \cite{TWED:Scholkopf01} and \cite{TWED:Shawe04} present a large synthesis of major results. We give hereinafter some basic definitions and some results derived from these previous references.

\begin{definition}
\label{Kernels}
A kernel on a non empty set $U$ refers to a complex (or real) valued symmetric function $\varphi(x,y) : U \times U \rightarrow \mathbb{C}$ (or $\mathbb{R}$).  
\end{definition}

\begin{definition}
\label{Definite Kernels}
Let $U$ be a non empty set. A function $\varphi: U \times U \rightarrow \mathbb{C}$ is called a positive (resp. negative) definite kernel if and only if it is Hermitian 
(i.e. $\varphi(x,y)=\overline{\varphi(y,x)}$ where the \textit{overline} stands for the conjugate number) for all $x$ and $y$ in $U$ and $\sum_{i,j=1}^n c_i \bar{c_j} \varphi(x_i, x_j) \ge 0$ (resp. $\sum_{i,j=1}^n c_i \bar{c_j} \varphi(x_i, x_j) \le 0$), 
for all $n$ in $\mathbb{N}$, $(x_1,x_2, ..., x_n) \in U^n$ and $(c_1,c_2,...,c_n) \in \mathbb{C}^n$.\\
\end{definition}

\begin{definition}
 \label{Conditionally Definite Kernels}

Let $U$ be a non empty set. A function $\varphi: U \times U \rightarrow \mathbb{C}$ is called a conditionally positive (resp. conditionally negative) definite kernel if and only if it is Hermitian 
(i.e. $\varphi(x,y)=\overline{\varphi(y,x)}$ for all $x$ and $y$ in $U$) and $\sum_{i,j=1}^n c_i \bar{c_j} \varphi(x_i, x_j) \ge 0$ (resp. $\sum_{i,j=1}^n c_i \bar{c_j} \varphi(x_i, x_j) \le 0$),
for all $n \ge 2$ in $\mathbb{N}$, $(x_1,x_2, ..., x_n) \in U^n$ and $(c_1,c_2,...,c_n) \in \mathbb{C}^n$ with $\sum_{i=1}^n c_i=0$. \\
\end{definition}

In the last two above definitions, it is easy to show that it is sufficient to consider mutually different elements in $U$, i.e. collections of distinct elements $x_1, x_2, ..., x_n$. This is what we will consider for the remaining of the paper.\\

\begin{definition}
 \label{matrix and kernel}
A positive (resp. negative) definite kernel defined on a finite set $U$ is also called a positive (resp. negative) semidefinite matrix. 
Similarly, a positive (resp. negative) conditionally definite kernel defined on a finite set is also called a positive (resp. negative) conditionally semidefinite matrix.
\end{definition}

For convenience sake, we will use p.d. and c.p.d. for positive definite and conditionally positive definite to characterize either a kernel or a matrix having these properties.\\

Constructing p.d. kernels from c.p.d. kernels is quite straightforward. For instance, if $-\varphi$ is a c.p.d. kernel on a set $U$ and $x_0 \in U$ then, according to \cite{TWED:BergChristensenRessel84}, $\psi(x,y) = \varphi(x,x_0)+\overline{\varphi(y, x_0)} - \varphi(x,y) - \varphi(x_0, x_0)$ is a p.d. kernel, so are $e^{(\psi(x,y))}$ and $e^{-\varphi(x,y)}$. The converse is also true.\\
Furthermore, $e^{-t \varphi(x,y)}$ is p.d. for $t>0$ if $-\varphi$ is c.p.d. We will precisely use this last result to construct p.d. kernels from classical elastic distances. 

\subsection{Sequence set}
\begin{definition}
 \label{Sequence set}
Let $\mathbb{U}$ be the set of finite sequences (symbolic sequences or time series): $\mathbb{U}$ $= \{A_{1}^{p} | p\in \mathbb{N}\}$. $A_{1}^{p}$ is a sequence with discrete index varying between $1$ and $p$. We note $\Omega$ the empty sequence (with null length) and by convention $A_{1}^{0}=\Omega$ so that $\Omega$ is a member of set $\mathbb{U}$. $|A|$ denotes the length of the sequence $A$.
Let $\mathbb{U}_p$ = $\{A \in \mathbb{U}\ | \  |A|\ \le p \}$ be the set of sequences whose length is shorter or equal to $p$.\\
\end{definition}

\begin{definition}
\label{Sequence element}
Let $A$ be a finite sequence. Let $A(i)$ be the $i^{th}$ element (symbol or sample) of sequence $A$. We will consider that $A(i) \in S \times T$ where $S$ embeds the multidimensional space variables (either symbolic or numeric) and $T \subset \mathbb{R}$ embeds the time stamp variable, so that we can write $A(i)=(a(i),t(i))$ where $a(i) \in S$ and $t(i) \in T$, with the condition that $t(i)> t(j)$ whenever $i>j$ (time stamps strictly increase in the sequence of samples).
$A_{i}^{j}$ with $i \leq j$ is the subsequence consisting of the $i_{th}$ through the $j_{th}$ element (inclusive) of $A$. So $A_{i}^{j}=A(i)A(i+1)...A(j)$. $A_{i}^{j}$ with $i>j$ is the null time series, e.g. $\Omega$. Furthermore, if $i>|A|$ then by convention, $A(i)=\Omega$.\\
\end{definition}

\subsection{General Edit/Elastic distance on a sequence set}

\begin{definition}
\label{Edit operations}
An edit operation is a pair $(a,b) \in ((S \times T) \cup \{\Omega\})^{2}$ written $a \rightarrow b$. The sequence $B$ results from the application of the edit operation $a \rightarrow b$ into sequence $A$, written $A \Rightarrow B$ via $a \rightarrow b$, if $A = \sigma a \tau$ and $B = \sigma b \tau$ for some sub-sequences $\sigma$ and $\tau$. We call $a \rightarrow b$ a substitution operation if $a \neq \Omega$ and $b \neq \Omega$, a delete operation if $b=\Omega$, an insert operation if $a=\Omega$. \\

For any pair of sequences $A_{1}^{p},B_{1}^{q}$, for which we consider the extensions $A_{0}^{p},B_{0}^{q}$ whose first element is $\Omega$,  and for each elementary edit operation related to position $0 \le i \le p$ in sequence $A$ and to position $0 \le j \le q$ in sequence $B$ is associated a cost value $\Gamma_{A(i) \rightarrow B(j)}(A_{1}^{p},B_{1}^{q})$, or $\Gamma_{A(i) \rightarrow \Omega,B,j}(A_{1}^{p},B_{1}^{q})$ or  $\Gamma_{\Omega,A,i \rightarrow B(j)}(A_{1}^{p},B_{1}^{q}) \in \mathbb{R}$.  To simplify this notation, we will simply write $\Gamma(A(i) \rightarrow B(j))$, $\Gamma(A(i) \rightarrow \Omega_B(j))$ or $\Gamma(\Omega_A(i) \rightarrow B(j))$ with the understanding that the suppression cost $\Gamma(A(i) \rightarrow \Omega_B(j))$ (respectively the insertion cost $\Gamma(\Omega_A(i) \rightarrow B(j))$) may depend on the current location $j$ in sequence $B$ (respectively on the location $i$ in sequence $A$). 

Hence we consider that $\Omega_X(k)$ is a function from $\mathbb{U}\times \mathbb{N}$ to $(S\times T) \cup \{\Omega\}$. \\


\end{definition}

\begin{definition}
\label{Edit distance}
A function $\delta: \mathbb{U} \times \mathbb{U} \rightarrow \mathbb{R}$ is called an edit distance defined on $\mathbb{U}$ if, for any pair of sequences $A_{1}^{p},B_{1}^{q}$, the following recursive equation is satisfied
\begin{eqnarray}
\label{Eq.1}
\begin{array}{ll}
 &\delta(A_{1}^{p},B_{1}^{q})=  \\
 & \text{Min} \left\{
   \begin{array}{ll}
     \delta(A_{1}^{p-1},B_{1}^{q})+\Gamma(A(p)\rightarrow \Omega_B(q)) & del \nonumber \\
     \delta(A_{1}^{p-1},B_{1}^{q-1})+\Gamma(A(p)\rightarrow B(q)) & sub \nonumber \\
     \delta(A_{1}^{p},B_{1}^{q-1})+\Gamma(\Omega_A(p) \rightarrow B(q)) & ins \nonumber \\
   \end{array}
   \right.
  \end{array}
\end{eqnarray}
\end{definition}

Note that not all edit/elastic distances are metric. In particular, the dynamic time warping distance does not satisfy the triangle inequality. 

\subsubsection{Levenshtein distance}
The Levenshtein distance $\delta_{lev}(x,y)$ has been defined for string matching. For this edit distance, the \textit{delete} and \textit{insert} operations induce unitary costs, i.e. $\Gamma(A(p)\rightarrow \Omega_B(q))=\Gamma(\Omega_A(p) \rightarrow B(q))=1$ while the \textit{substitution} cost is null if $A(p)=B(q)$ or $1$ otherwise. Thus, for this distance, we consider that the functional term $\Omega_X(k)=\Omega$ for all $X$ and $k$, since the suppression and insertion costs do not depend on the suppressed or inserted element respectively.

\subsubsection{Dynamic time warping}
The DTW similarity measure $\delta_{dtw}$ \cite{TWED:Velichko70}\cite{TWED:Sakoe71} is defined according to the previous notations such as:

\begin{eqnarray}
\label{Eq.2}
 \delta_{dtw}(A_{1}^{p},B_{1}^{q})&= &d_{LP}(a(p),b(q))\nonumber \\
  &+&\text{Min}
   \left\{
   \begin{array}{ll}
     \delta_{dtw}(A_{1}^{p-1},B_{1}^{q}) & del \nonumber \\
     \delta_{dtw}(A_{1}^{p-1},B_{1}^{q-1}) & sub\nonumber \\
     \delta_{dtw}(A_{1}^{p},B_{1}^{q-1}) & ins\nonumber \\
   \end{array}
   \right.
\end{eqnarray}

where  and $d_{LP}(x,y)$ is the $LP$ norm of vector $(x-y)$ in $S$,
and so for DTW, $\Gamma(A(p)\rightarrow \Omega_B(q))=\Gamma(A(p)\rightarrow B(q))=\Gamma(\Omega_A(p) \rightarrow B(q))=d_{LP}(a(p),b(q))$. Thus, for this distance, $\Omega_X(k)=X(k)$ for all $X$ and $k$.
Let us note that the time stamp values are not used, therefore the costs of each edit operation involve vectors $a$ and $b$ in $S$ instead of vectors $(a, t_a)$ and $(b, t_b)$ in $S \times T$. Furthermore, $\delta_{dtw}$ does not comply with the triangle inequality as shown in \cite{TWED:Chen04}.

\subsubsection{Edit Distance with real penalty}

\begin{eqnarray}
\begin{array}{ll}
\label{Eq.3}
 \delta_{erp}(A_{1}^{p},B_{1}^{q})= \\
 \text{Min} \left\{
   \begin{array}{ll}
     \delta_{erp}(A_{1}^{p-1},B_{1}^{q})+\Gamma(A(p)\rightarrow \Omega_B(q)) & del  \nonumber\\
     \delta_{erp}(A_{1}^{p-1},B_{1}^{q-1})+\Gamma(A(p)\rightarrow B(q))  & sub \nonumber\\
     \delta_{erp}(A_{1}^{p},B_{1}^{q-1})+\Gamma(\Omega_A(p) \rightarrow B(q))& ins  \nonumber\\
   \end{array}
   \right.
 \end{array}
\end{eqnarray}
with
\begin{eqnarray}
\label{erp} 
   \begin{array}{ll}
     \Gamma(A(p)\rightarrow \Omega_B(q))=d_{LP}(a(p),g))  \nonumber \\
     \Gamma(A(p)\rightarrow B(q)=d_{LP}(a(p),b(q))  \nonumber \\
     \Gamma(\Omega_A(p) \rightarrow B(q))=d_{LP}(g,b(q))  \nonumber \\
   \end{array}
\end{eqnarray}
where $g$ is a constant in $S$. For this distance, $\Omega_X(k)=g$ for all $X$ and $k$.

Note that the time stamp coordinate is not taken into account, therefore $\delta_{erp}$ is a distance on $S$ but not on $S \times T$. Thus, the cost of each edit operation involves vectors $a$ and $b$ in $\mathbb{R}^k$ instead of vectors $(a,t_a)$ and $(b, t_n)$ in $\mathbb{R}^{k+1}$.

According to the authors of ERP \cite{TWED:Chen04}, the constant $g$ should be set to $0$ for some intuitive geometric interpretation and in order to preserve the mean value of the transformed time series when adding gap samples.

\subsubsection{Time warp edit distance}

Time Warp Edit Distance (TWED) \cite{TWED:Marteau08}, \cite{TWED:Marteau09} is defined similarly to the edit distance defined for string \cite{TWED:Levenshtein66}\cite{TWED:Wagner73}. The similarity between any two time series $A$ and $B$ of finite length, respectively $p$ and $q$ is defined as:

$\delta_{twed}(A_{1}^{p},B_{1}^{q})=$
\begin{eqnarray}
\label{twed}
 \text{Min} \left\{
   \begin{array}{ll}
     \delta_{twed}(A_{1}^{p-1},B_{1}^{q})+\Gamma(A(p)\rightarrow \Omega_B(q)) & del_A \nonumber \\
     \delta_{twed}(A_{1}^{p-1},B_{1}^{q-1})+\Gamma(A(p)\rightarrow B(q)) & subs \nonumber \\
     \delta_{twed}(A_{1}^{p},B_{1}^{q-1})+\Gamma(\Omega_A(p) \rightarrow B(q)) & del_B \nonumber \\
   \end{array}
   \right.
\end{eqnarray}
with
\begin{eqnarray}
\label{twed_b} 
   \begin{array}{ll}
     \Gamma(A(p)\rightarrow \Omega_B(q))=d(A(p),A(p-1))+\lambda  \nonumber \\
     \Gamma(A(p)\rightarrow B(q))=d(A(p),B(q))+d(A(p-1),B(q-1)) \nonumber \\
     \Gamma(\Omega_A(p) \rightarrow B(q))=d(B(q),B(q-1))+\lambda  \nonumber \\
   \end{array}
\end{eqnarray}
where $\lambda$ is a positive constant that represents a gap penalty. The time stamps are exploited to evaluate $d(A(p),B(q))$, in practice, $d(A(p),B(q))=d_{LP}(a(p),b(q)) + \nu  d_{LP}(t(p), t(q))$, where $\nu$ is a non negative constant which characterizes the \textit{stiffness} of the $\delta_{twed}$ elastic measure along the \textit{time} axis.
 Furthermore, for this distance, we consider that the functional term $\Omega_X(k)=\Omega$ for all $X$ and $k$, since the suppression and insertion costs do not depend on the suppressed or inserted element respectively.

\subsection{Indefiniteness of elastic distance kernels}

In appendix \ref{Appendix:A}, we give counter examples, one for each previously defined elastic distance, showing that these distances do not lead to definite kernels. This demonstrates that the metric properties of a distance defined on $\mathbb{U}$, in particular the triangle inequality, are not sufficient conditions to establish definiteness (conditionally or not) of the associated distance kernel. One could conjecture that elastic distances cannot be definite (conditionally or not), possibly because of the presence of the $\max$ or $\min$ operators in the recursive equation. In the following sections, we will see that replacing these $\min$ or $\max$ operators by a sum operator makes possible, under some conditions, for the construction of series of positive definite kernels whose limit is quite directly connected to the previously addressed elastic distance kernels.

\section{Constructing positive definite kernels from elastic distance}

The main idea leading to the construction of positive definite kernels from a given elastic distance defined on $\mathbb{U}$ is to replace the $\min$ or $\max$ operator into the recursive equation defining the elastic distance by a summation ($\sum$) operator. Instead of keeping only one of the best alignment paths, the new kernel will sum up the costs of all the existing sub-sequence alignment paths with some weighting factor that will favor \textit{good} alignments while penalizing bad alignments. In addition, this weighting factor can be optimized. This principle has been applied successfully to the Smith and Waterman symbolic distance which is also known to be indefinite \cite{TWED:Saigo04}, and more recently to dynamic time warping kernels for time series alignment \cite{TWED:Cuturi07}. If, basically, we are following the same objective, the approach that we propose is quite different since it is based on a regularization principle which is simply and directly expressed into the recursive equation defining the elastic distance. First we replace the \textit{min} (or \textit{max}) operator by a summation operator and introduce a symmetric \textit{corridor} function to optionally limit the summation. Then we add a regularizing recursive term such that the proof of the positive definiteness property  can be understood as a direct consequence of the Haussler's convolution theorem, as shown in appendix \ref{Appendix:B}. Basically, all is reduced to the proper inventory of pairs of symmetric alignment paths.
 
\subsection{Recursive accumulation of \textit{f-cost} products}

\begin{definition}
\label{RAfP}
A function $\mathcal{C}(.,.): \mathbb{U} \times \mathbb{U} \rightarrow \mathbb{R}$ is called a Recursive Accumulation of \textit{f-cost} Products (RAfP) if, for any pair of sequences $A_{1}^{p},B_{1}^{q}$, there exist a function $f: \mathbb{R} \rightarrow \mathbb{R}$ and three indicator functions $h_d$, $h_s$ and $h_i: \mathbb{N}^{2} \rightarrow \{0,1\}$ such that the following recursive equation is satisfied:
\begin{align}
\label{Eq.4}
 &\mathcal{C}(A_{1}^{p},B_{1}^{q})= \frac{1}{c}\\
 & \sum \left\{
   \begin{array}{ll}
     h_d(p,q) \mathcal{C}(A_{1}^{p-1},B_{1}^{q}) f(\Gamma_d(A, p, B, q)) & del \notag\\
     h_s(p,q) \mathcal{C}(A_{1}^{p-1},B_{1}^{q-1}) f(\Gamma_s(A, p, B, q)) & sub \notag\\
     h_i(p,q) \mathcal{C}(A_{1}^{p},B_{1}^{q-1}) f(\Gamma_i(A, p, B, q)) & ins \notag
   \end{array}
   \right.
\end{align}
\end{definition}

Where $c$ is a non negative constant. This recursive definition relies on an initialization. To that end we set $\mathcal{C}(\Omega,\Omega)=\xi$, where $\xi$ is a real non negative constant, typically $\xi=1$, and $\Omega$ the null sequence. Note that if no path satisfying the indicator functions exists when aligning time series $A$ and $B$, then $\mathcal{C}(A,B)=0$. 

Furthermore, according to the indicator functions $h_d$, $h_s$ and $h_i$, this type of construction sums up the multiplication of the local quantities  $f(\Gamma(A(i) \rightarrow B(j)))$ that we call \textit{f-costs} for all or for some of the possible alignment paths that exist between the two time series, the concept of alignment path being precisely defined in appendix \ref{Appendix:B} (see definition \ref{def:piMap}).  

In this paper we are specifically concerned with the following two symmetric RAfP families. The first one (Eq. \ref{Eq.RAfP_h}), characterized by $\forall (p,q) \in \mathbb{N}^{2}$, $h_d(p,q) = h(p-1,q)$, $ h_i(q,p) = h(p,q-1)$ and $h_s(p,q) = h(p-1, q-1)$, accumulates the  \textit{f-cost} products along all the alignment paths between the two time series (Eq.\ref{Eq.RAfP_h}) that exist inside the symmetric \textit{corridor} defined by $h(p,q)$. For the case depicted in Figure \ref{fig:DTW_Corridor}, $h(p,q)=1$, whenever $|p-q| \le 3$, $h(p,q)=0$ otherwise.  We denote this RAfP function $\mathcal{C}_{h}$. 

\begin{align}
 &\mathcal{C}_h(A_{1}^{p},B_{1}^{q})= \frac{1}{c}\label{Eq.RAfP_h}\\
 & \sum \left\{
   \begin{array}{ll}
    h(p-1,q) \mathcal{C}_h(A_{1}^{p-1},B_{1}^{q}) f(\Gamma(A(p)\rightarrow \Omega_B(q)))  \notag\\
    h(p-1,q-1) \mathcal{C}_h(A_{1}^{p-1},B_{1}^{q-1})f(\Gamma(A(p)\rightarrow B(q))) \notag\\
    h(p,q-1) \mathcal{C}_h(A_{1}^{p},B_{1}^{q-1}) f(\Gamma(\Omega_A(p) \rightarrow B(q))) \notag
   \end{array}
   \right.
\end{align}

The second RAfP function (Eq. \ref{Eq.RAfP_Kh}) accumulates the \textit{f-cost} products of all mappings characterized by simultaneous insertions, deletions and substitutions. Hence the substitution, deletion and insertion operations are simultaneously performed on the same index value. This function is thus characterized by $\forall (p,q) \in \mathbb{N}^{2}$, $h_d(p,q) = h(p-1,q)$, $h_i(q,p) = h(p,q-1)$ where $h$ is a symmetric function, and $h_s(p,q) = \delta_{p,q}h(p-1,q-1)$, where $\delta$ is the Kronecker's symbol (Eq.\ref{Eq.RAfP_Kh}). We denote this second RAfP function $\mathcal{C}_{h,\delta}$. 

\begin{align}
 &\mathcal{C}_{h,\delta}(A_{1}^{p},B_{1}^{q})=  \frac{1}{c}\label{Eq.RAfP_Kh}\\
 & \sum \left\{
   \begin{array}{ll}
    h(p-1,q) \mathcal{C}_{h,\delta}(A_{1}^{p-1},B_{1}^{q}) f(\Gamma(\Omega_A(p)\rightarrow \Omega_B(p))  \\
    \delta_{p,q}h(p-1,q-1)\cdot \\
    \hspace{1cm}\mathcal{C}_{h,\delta}(A_{1}^{p-1},B_{1}^{q-1})f(\Gamma(A(p)\rightarrow B(q))) \notag \\
   h(p,q-1) \mathcal{C}_{h,\delta}(A_{1}^{p},B_{1}^{q-1}) f(\Gamma(A(q) \rightarrow B(q))   \\
   \end{array}
   \right.
\end{align}

 
\subsection{Recursive Edit Distance Kernel - Main result}
\begin{definition}
\label{REDK}
We call Recursive Edit Distance Kernel (REDK) the function $\mathcal{K}(.,.)=\mathcal{C}_{h}(.,.)+\mathcal{C}_{h,\delta}(.,.):$ $\mathbb{U} \times \mathbb{U} \rightarrow \mathbb{R}$.  
\end{definition}

The following theorem, that relates to the ones presented in \cite{TWED:Cuturi07} and \cite{TWED:ShinK08}, establishes sufficient conditions on $f(\Gamma(a \rightarrow b))$ for a REDK function to be definite positive, and thus is a basis for the construction of definite REDK. The proof of this theorem, that is given in Appendix \ref{Appendix:B}, is a direct consequence of the Haussler’s \textit{R-convolution} theorem \cite{TWED:Haussler99}. The sufficient condition that is proposed here is quite general and less restrictive than those proposed by \cite{TWED:Cuturi07} and \cite{TWED:ShinK08}. The avenue that is taken here only requires that the local kernel $k(x,y)$ is PD. In previous works, the condition for positive definiteness is that  the local kernel $k(x,y)$ and the ratio $k(x,y)/(k(x,y)+1)$ are jointly PD. Our new condition is obviously weaker than the one proposed in [9]. Furthermore, our REDK construction applies to a wider range of edit or time warp distances developed either for symbolic sequence or time series matching. \\

\begin{theorem}
\label{theorem:Definiteness of REDK}
Definiteness of REDK:\\

REDK is a positive definite kernel on $\mathbb{U}\times \mathbb{U}$ iff the local kernel $k(x,y)=f(\Gamma(x \rightarrow y))$ is a positive definite kernel on $\left((S \times T) \cup \{\Omega\}\right)^2$.\\
\end{theorem}

A sketch of proof for theorem \ref{theorem:Definiteness of REDK} is given in the appendix \ref{Appendix:B}.\\

As the cost function  $\Gamma$ is, in general, conditionally negative definite, choosing for $f(.)$ the exponential ensures that $f(\Gamma(x \rightarrow y))$ is a positive definite kernel \cite{TWED:Schoenberg38}.  

Other functions can be used, such as the Inverse Multi Quadric kernel $k(x, y) = \frac{1}{\sqrt{\left( \Gamma(x \rightarrow y) \right)^2 + \theta^2}}$. As with the exponential (Gaussian or Laplace) kernel, the Inverse Multi Quadric kernel results in a positive definite matrix with full rank \cite{TWED:Micheli86} and thus forms an infinite dimension feature space. Note here that Cuturi et al. \cite{TWED:Cuturi07} state that to ensure the definiteness of the DTW-REDK kernel, not only $f(\Gamma(x \rightarrow y))$, but also $f(\Gamma(x \rightarrow y))/(1+f(\Gamma(x \rightarrow y)))$ need to be PD kernels, which forms a stronger condition than the one we propose.

\subsubsection{Computational cost of REDK}
\begin{figure}[]
\centering
\includegraphics[scale=0.45]{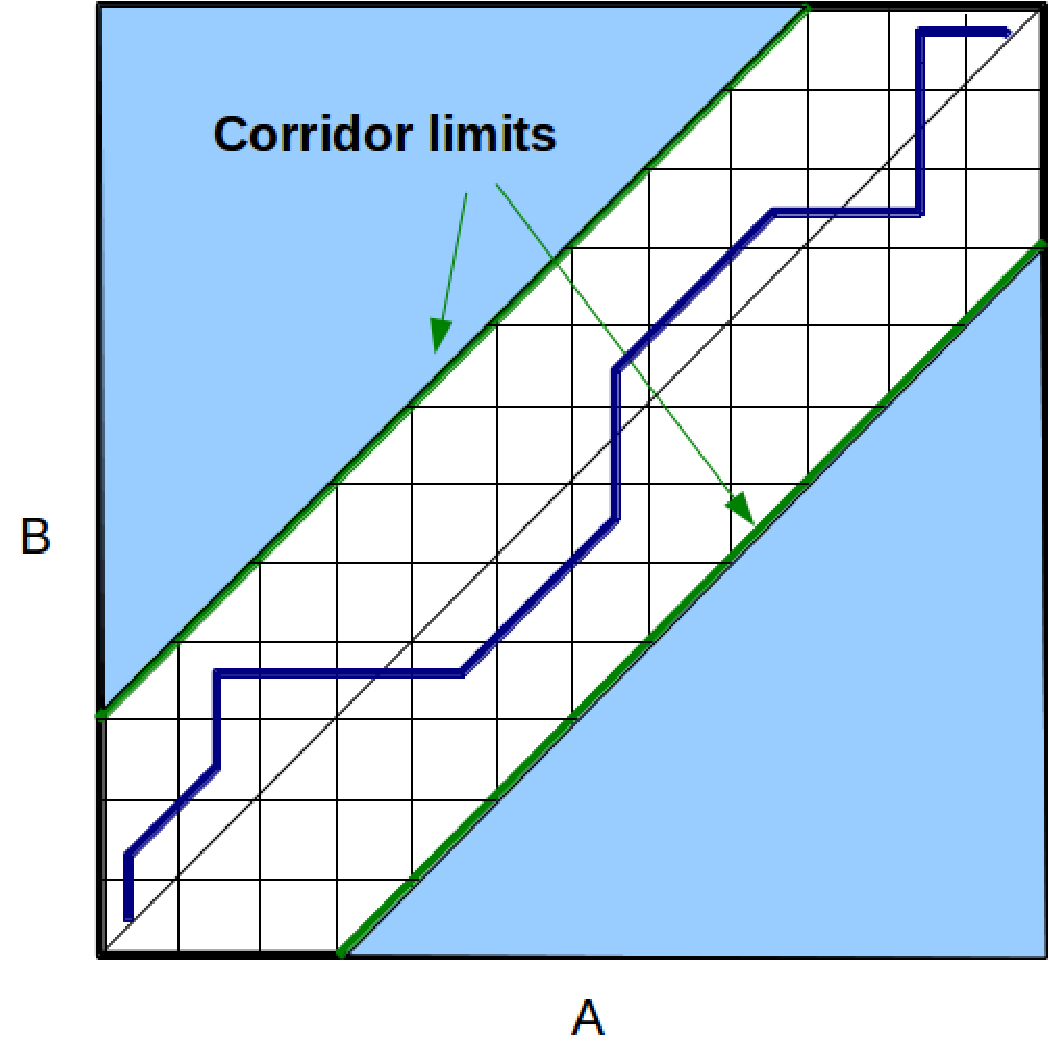}
\caption{Example of alignment paths existing when matching time series $A$ and $B$. In red dotted line a path whose editing \textit{f-cost} product is accumulated by $\mathcal{C}_{h}$ and $\mathcal{C}_{h,\delta}$, in blue plain line, a path whose editing cost product is accumulated by $\mathcal{C}_{h}$ but not $\mathcal{C}_{h,\delta}$. A symmetric corridor of any form can be used to reduce the number of admissible alignment paths that are inventoried by a RAfP}
\label{fig:DTW_Corridor}
\end{figure}

Although the number of paths that are summed up exponentially increases with the lengths $|A|$ and $|B|$ of the time series that are evaluated, the recursive computation of any RAfP function and therefore of $\mathcal{K}(A,B)$ lead to a quadratic computational cost $O(|A||B|)$, e.g. $O(N^2)$ if $N$ is the average length of the time series that are considered. This quadratic complexity can be reduced to a linear complexity by limiting the number of alignment paths to consider in the recursion. This can be achieved when using a search corridor \cite{TWED:Sakoe71} as far as the kernel remains symmetric, which is the case when processing time series of equal lengths and restraining the search space using, for instance, a fixed size corridor symmetrically displayed around the diagonal as shown in Fig. \ref{fig:DTW_Corridor}. Note that any kind of corridor can be specified using the indicator functions $h_d$, $h_s$ and $h_i$. 

\subsection{\textit{Exponentiated} REDK}

\begin{definition}
The following equations define the two RAfP function instances entering into the construction of the exponentiated REDK, considering $f(\Gamma(. \rightarrow .))=\frac{1}{c}.e^{-\nu' \Gamma(. \rightarrow .)}$ with $c>0$

\begin{align}
 &\mathcal{C}_{e,h}(A_{1}^{p}, B_{1}^{q})= \frac{1}{c} \label{Eq.MEREDK}\\
 & \sum \left\{
   \begin{array}{ll}
    h(p-1,q) \mathcal{C}_{e,h}(A_{1}^{p-1},B_{1}^{q})\hspace{1mm}e^{-\nu' \Gamma(A(p)\rightarrow \Omega_B(q))} \\
    h(p-1,q-1) \mathcal{C}_{e,h}(A_{1}^{p-1},B_{1}^{q-1})\hspace{1mm} e^{-\nu' \Gamma(A(p)\rightarrow B(q))}  \nonumber\\
    h(p,q-1) \mathcal{C}_{e,h}(A_{1}^{p},B_{1}^{q-1})\hspace{1mm} e^{-\nu' \Gamma(\Omega_A(p) \rightarrow B(q))} \\
   \end{array}
   \right.\end{align}

\begin{align}
 &\mathcal{C}_{e, h, \delta}(A_{1}^{p}, B_{1}^{q})= \frac{1}{c} \label{Eq.MEREDK-d}\\
 & \sum \left\{
   \begin{array}{ll}
    h(p-1,q) \mathcal{C}_{e, h, \delta}(A_{1}^{p-1},B_{1}^{q}) e^{-\nu' \Gamma(\Omega_A(p)\rightarrow \Omega_B(p))} \\
    \delta_{pq} h(p-1,q-1) \mathcal{C}_{e, h, \delta}(A_{1}^{p-1},B_{1}^{q-1}) e^{-\nu' \Gamma(A(p)\rightarrow B(q))} \\
    h(p,q-1) \mathcal{C}_{e, h, \delta}(A_{1}^{p},B_{1}^{q-1}) e^{-\nu' \Gamma(A(q)\rightarrow B(q))} \\ \notag
   \end{array}
   \right.
\end{align}

\end{definition}

where $\nu'$ is a \textit{stiffness} parameter that weights the contribution of the local elementary costs. The larger $\nu'$ is, the more the kernel is selective around the optimal paths. At the limit, when $\nu' \rightarrow \infty$, only the optimal path costs are summed up by the kernel. Note that, as is generally seen, several optimal paths leading to the same global cost exist, therefore $\lim_{\nu' \to +\infty} -1/\nu'  log(\mathcal{K}_{e}(A,B)) $ does not coincide with the elastic distance $\delta$ that involves the same corresponding elementary costs, although we expect it to be close. 

The constant $1/c$ (typically $1/3$) is used to maintain the global accumulation functions $\mathcal{C}_{e,h}$ and $\mathcal{C}_{e, h, \delta}$ in a range that is computationally acceptable. 

The alignment cost of two null time series being $0$, we set $\mathcal{C}_{e, h}(\Omega, \Omega) = \mathcal{C}_{e, h, \delta}(\Omega, \Omega)=1$ in this exponentiation context.\\

\begin{proposition}
\label{DefinitenesREDKexp}
Definiteness of the exponentiated REDK instance, $\mathcal{K}_{e}$:\\
$\mathcal{K}_{e}(.,.)$ is positive definite for the cost functions $\Gamma(A(p)\rightarrow \Omega_B(q))$, $\Gamma(A(p)\rightarrow B(q))$ and $\Gamma(\Omega_A(p) \rightarrow B(q))$ involved in the computation of the $\delta_{lev}$, $\delta_{dtw}$ and $\delta_{erp}$ distances.\\
\end{proposition}
The proof of proposition \ref{DefinitenesREDKexp} is straightforward and is omitted.\\

The local cost function $\Gamma(\Omega_A(p) \rightarrow B(q))$ involved in the computation of the $\delta_{twed}$ does not respect the condition of theorem \ref{theorem:Definiteness of REDK} and thus the local kernel is not guaranteed yet to be p.d. for this distance.\\
 
The  REDKs constructed from these distances are referred respectively to $REDK_{lev}$, $REDK_{erp}$, $REDK_{dtw}$, $REDK_{twed}$ in the rest of the paper.\\

\subsubsection{Interpretation of the exponentiated REDK}

In a REDK kernel, each alignment path is assigned with a product of \textit{f-costs} that is the multiplication of the local \textit{f-cost} functions attached to each edge of the path. For exponentiated REDK, the local \textit{f-cost} function, e.g. $e^{-\nu' \Gamma(A(p)\rightarrow B(q))}$ can be interpreted, up to a normalizing constant, as a probability to align symbol $A(p)$ with symbol $B(q)$, and the value affected to each path can be interpreted as the probability of a specific alignment between two sequences. In this case the REDK, that sums up the probability of all possible alignment paths between two sequences, can be interpreted as a matching probability between two sequences. This probabilistic interpretation suggests an analogy between REDK and the \textit{alpha-beta} algorithm designed to learn HMM models: while the Viterbi's algorithm that uses a \textit{max} operator in a dynamic programming implementation (just like the \textit{DTW} algorithm) evaluates only the probability of the best alignment path, the \textit{alpha-beta} algorithm is based on the summation of the probabilities of all possible alignment paths. As reported in \cite{TWED:Saigo04}, the main drawbacks of these kind of kernels is the vanishing of the product of local cost functions (that are lower than one) when comparing long sequences. When considering distance-matrix (obtained from pairwise distances on finite sets) this leads to a matrix that suffers from the diagonal dominance problem, i.e. the fact that the kernel value decreases extremely fast when the similarity slightly decreases. \\

\section{Classification experiments}

The purpose of this experiment is to estimate the benefit one can expect from using PD elastic kernels instead of indefinite ones in SVM classification. It is clear that the the Sequential Minimal Optimization (SMO) involved in the SVM learning procedure \cite{Platt1999} \cite{P-H.Chen2006} is able to handle indefinite kernels, but, as the convergence toward a global extremum is not guaranteed (the optimization is problem is not convex), we can expect some better accuracy for definite kernels especially when the pairwise distance matrix derived from the train data set is far from definiteness. This is precisely what our experiment targets to demonstrate.

To that end, we empirically evaluate the effectiveness of some REDK kernels comparatively to Gaussian Radial Basis Function (RBF) Kernels or elastic distance substituting kernels \cite{TWED:Haasdonk04} using some classification tasks on a set of time series coming from quite different application fields.  The classification task we have considered consists of assigning one of the possible categories to an unknown time series for the 20 data sets available at the UCR repository \cite{TWED:KeoghUCRdataset}. As time is not explicitly given for these datasets, we used the index value of the samples as the time stamps for the whole experiment.\\

For each dataset, a training subset (TRAIN) is defined as well as an independent testing subset (TEST). We use the training sets to train two kinds of classifiers:
\begin{itemize}
 \item the first one is a first near neighbor (1-NN) classifier: first we select a training data set containing time series for which the correct category is known. To assign a category to an unknown time series selected from a testing data set (different from the train set), we select its nearest neighbor (in the sense of a distance or similarity measure) within the training data set, then, assign the associated category to its nearest neighbor. For that experiment, a leave one out procedure is performed on the training dataset to optimized the meta parameters of the considered comparability measure.
 \item the second one is a SVM classifier \cite{TWED:BoserGV92}, \cite{TWED:Vapnik1995} configured with a Gaussian RBF kernel whose parameters are $C > 0$, a trade-off between regularization and constraint violation and $\sigma$ that determines the width of the Gaussian function. To determine the $C$ and $\sigma$ hyper parameter values, we adopt a 10-folded cross-validation method on each training subset. According to this procedure, given a predefined training set TRAIN and a test set TEST, we adapt the meta parameters based on the training set TRAIN: we first divide TRAIN into 10 stratified subsets ${TRAIN_1, TRAIN_2,\cdots, TRAIN_{10}}$; then for each subset $TRAIN_i$ we use it as a new test set, and regard $(TRAIN-TRAIN_i)$ as a new training set; Based on the average error rate obtained on the 10 classification tasks, the optimal values of meta parameters are selected as the ones leading to the minimal
average error rate.\\
\end{itemize}

We have used the LIBSVM library \cite{libsvm01} to implement the SVM classifiers.

\subsection{Experimenting with REDK}

We tested the exponentiated REDK kernel based on the $\delta_{erp}, \delta_{dtw}, \delta_{twed}$ distance costs. We consider respectively the positive definite $REDK_{erp}, REDK_{dtw}, REDK_{twed}$ kernels. Our experiment compares classification errors on the test data for 
\begin{itemize}
 \item the  first near neighbor classifiers based on the $\delta_{erp}, \delta_{dtw}, \delta_{twed}$ distance measures (1-NN $\delta_{erp}$, 1-NN $\delta_{dtw}$ and 1-NN $\delta_{twed}$), 
 \item the SVM classifiers using Gaussian distance substituting kernels based on the same distances \cite{Zhang:2010}, e.g. SVM $\delta_{erp}$,  SVM $\delta_{dtw}$, SVM $\delta_{twed}$, 
 \item and their corresponding REDK kernel,  SVM $REDK_{erp}$, SVM $REDK_{dtw}$, SVM $REDK_{twed}$.
\end{itemize}

For $\delta_{erp}$, $\delta_{twed}$, $REDK_{erp}$ and $REDK_{twed}$  we used the L1-norm, while the L2-norm has been implemented for $\delta_{dtw}$ and $REDK_{dtw}$, a classical choice for DTW \cite{TWED:Ratanamahatana04}.

\subsubsection{Meta parameters}
For $\delta_{erp}$ kernel, meta parameter $g$ is optimized for each dataset on the train data by minimizing the classification error rate of a first near neighbor classifier using a Leave One Out (LOO) procedure. For this kernel, $g$ is selected in $\{-3, -2.99, -2.98, \cdots, 2.98, 2.99, 3\}$. This optimized value is also used for comparison with the $REDK_{erp}$ kernel.

For $\delta_{twed}$ kernel, meta parameters $\lambda$ and $\nu$ are optimized for each dataset on the train data by minimizing the classification error rate of a first near neighbor classifier. For our experiment, the \textit{stiffness} value ($\nu$) is selected from $\{10^{-5},  10^{-4}, 10^{-3}, 10^{-2}, 10^{-1}, 1\}$ and $\lambda$ is selected from $\{0, .25, .5, .75, 1.0\}$. If different $(\nu, \lambda)$ values lead to the minimal error rate estimated for the training data then the pairs containing the highest $\nu$ value are selected first, then the pair with the highest $\lambda$ value is finally selected.
These optimized ($\lambda$, $\nu$) values are also used for comparability purposes with the $REDK_{twed}$ kernel.

The kernels exploited by the SVM classifiers are the Gaussian Radial Basis Function (RBF) kernels  $K(A,B)=e^{\delta(A,B)^2/(2 \sigma^2)}$ where $\delta$ stands for $\delta_{erp}$, $\delta_{dtw}$, $\delta_{twed}$, $REDK_{erp}$, $REDK_{dtw}$, $REDK_{twed}$. To limit the search space for the SVM meta parameters, the pairwise distances or kernels values are normalized using the \textit{min} and \textit{max} values calculated on the TRAIN data, basically\\
\begin{itemize}
\item $\delta=(\delta-\delta_{min})/(\delta_{max}-\delta_{min})$ for $\delta = \delta_{erp}$, $\delta_{dtw}$ or $\delta_{twed}$, and
\item $\delta=exp\big((log(\delta)-log(\delta_{min}))/(log(\delta_{max})-log(\delta_{min}))\big)$ for $\delta = REDK_{erp}$, $REDK_{dtw}$, $REDK_{twed}$.\\
\end{itemize}   

Meta parameter $C$ is selected from $\{2^{-5}, 2^{-4}, ..., 1,2, ..., 2^{10}\}$, and $\sigma^2$ from $\{2^{-5}, 2^{-4}, ..., 1,2, ..., 2^{14}\}$. The best values are obtained using the 10-folds cross validation procedure. \\

For the $REDK_{erp}$, $REDK_{dtw}$ and $REDK_{twed}$ kernels, meta parameter $1/\nu'$ is selected from the discrete set $\mathcal{G}=\{10^{-5}, 10^{-4}, ..., 1,10,100\}$. 

The optimization procedure is as follows: 
\begin{itemize}
 \item for each value in $\mathcal{G}$, we train a SVM $REDK_{*}$ classifier on the training dataset using the previously described 10-folded cross validation procedure to select the SVM meta parameters $C$ and $\sigma$ and the average of the classification error is recorded.
 \item  the best $\sigma, C$ and $\nu'$ values are the ones that lead to the minimal average error.\\
\end{itemize}

Table \ref{Tab2} gives for each data set and each tested kernel ($\delta_{erp}$, $\delta_{dtw}$, $\delta_{twed}$, $REDK_{erp}$, $REDK_{dtw}$ and $REDK_{twed}$) the corresponding optimized values of the meta parameters. \\

\begin{table*}[ht]
 \caption{Meta parameters used in conjunction with $\delta_{erp}$, ${REDK}_{erp}$,  $\delta_{dtw}$, ${REDK}_{dtw}$, $\delta_{twed}$ and ${REDK}_{twed}$ kernels}
\label{Tab2}
\centering
\begin{tabular}{|l|c|c|c|c|c|c|c|}
\hline
\textbf{DATASET}  &  \textbf{$\delta_{erp}: g;C;\sigma$}  &  \textbf{${REDK}_{erp}: g;\nu';C;\sigma$} &  \textbf{$\delta_{dtw}: C;\sigma$} &  \textbf{${REDK}_{dtw}: \nu';C;\sigma$} & \textbf{$\delta_{twed}: \Omega;\nu;C;\sigma$} & \textbf{${REDK}_{twed}: \Omega;\nu;\nu';C;\sigma$} \\
\hline\hline
\textbf{Synth. cont.}  & 0.0;2.0;0.25 & 0.0;0.457;256.0;0.062   & 8.0;4.0 & 0.047;1024.0;0.062  & 0.75;0.01;1.0;0.25 & 0.75;0.01;0.685;8.0;4.0 \\
\hline
\textbf{Gun-Point} & -0.35;4.0;0.031  & -0.35;0.457;128.0;1.0 & 16.0;0.0312 & 0.457;64.0;2.0 & 0.0;0.001;8.0;1.0 & 0.0;0.001;0.685;32;32 \\
\hline
\textbf{CBF} &  -0.11;1.0;1.0 & -0.11;0.203;4.0;32.0  & 1.0;1.0 & 0.457;2.0;1.0 & 1.0;0.001;1.0;1.0 & 1.0;0.00;,0.20;4.0;32.0 \\
\hline
\textbf{Face (all)} &  -1.96;4.0;0.5 & -1.96;1.028;8.0;0.62 & 2.0;0.25 & 1.028;4.0;0.25 & 1.0;0.01;8.0;4.0 & 1.0;0.01;2.312;8.0;4.0  \\
\hline
\textbf{OSU Leaf} &  -2.25;2.0;0.062 & -2.25;1.541;256;0.031 & 4.0;0.062 & 1.541;32.0;0.062 & 1.0;1e-4;8.0;0.25 & 1.0;1e-4;1.028;64.0;1.0 \\
\hline
\textbf{Swed. Leaf}  & 0.3;8.0;0.125 & 0.3;0.203;1.0;4.0 & 4.0;0.031 & 5.202;0.062;0.5 & 1.0;1e-4;16.0;0.062 & 1.0;1e-4;0.304;32.0;1.0 \\
\hline
\textbf{50 Words}  & -1.39;16.0;0.25 & -1.39;0.685;16;0.25 & 4.0;0.062 & 1.028;64.0;0.062 & 1.0;1e-3;8.0;0.5 & 1.0;1e-3;1.028;32.0;2.0 \\
\hline
\textbf{Trace}  & 0.57;32;0.62 & 0.57;0.457;256;4.0 & 4;0.25 & 0.685;16;0.25 & 0.25;1e-3;8.0;0.25 & 0.25;1e-3;300;0.0625;0.25 \\
\hline
\textbf{Two Patt.}  & -0.89;0.25;0.125 & -0.89;0.304;0.004;1.0 & 0.25,0.125  & 0.457;2.0;0.125 & 1.0;1e-3;0.25;0.125 & 1.0;1e-3;0.685;0.25;0.125 \\
\hline
\textbf{Wafer} & 1.23;2.0;0.062 & 1.23;0.685;4.0;0.5 & 1.0;0.016 & 1.541;1024;0.031 & 1.0;0.125;4.0;0.62 & 1.0;0.125;1.541;1.0;4.0 \\
\hline
\textbf{face (four)}  & 1.97;64;16 & 1.97;0.685;32;2 & 16;0.5 & 0.457;16;2 & 1.0;0.01;4;2 & 1.0;0.01;1.027;4;2 \\
\hline
\textbf{Ligthing2}  & -0.33;2;0.062 & -0.33;2.312;128;0.062 & 2.0;0.031 & 1.541;32;0.062 & 0.0;1e-6;8;0.25 & 0.0;1e-6;1.541;8;8 \\
\hline
\textbf{Ligthing7} & -0.40;128;2 & -0.40;0.685;32;0.25 & 4;0.25 & 0.685;32;0.062 & 0.25;01;4;0.5 & 0.25;0.1;0.685;4;8 \\
\hline
\textbf{ECG}  & 1.75;8;0.125 & 1.75;0.457;16;0.5 & 2;0.62 & 1.028;32;0.062 &0.5;1.0;4;0.125 & 0.5;1.0;5.202;8;16 \\
\hline
\textbf{Adiac} & 1.83;16;0.0156 & 1.83;2.312;4096;0.031 & 16;0.0039 & 1.028;2048;0.031 & 0.75;1e-4;16;0.016 & 0.75;1e-4;2.312;128;1 \\
\hline
\textbf{Yoga} & 0.77;4;0.031 & 0.77;11.7054096;0.031 & 4;0.008 & 26.337;1024;0.031 & 0.5;1e-5;2;0.125 & 0.5;1e-5;3.468;256;2 \\
\hline
\textbf{Fish}  & -0.82;64;0.25 & -0.82;0.685;32;0.5 & 8;0.016 & 3.468;64;16 & 0.5;1e-4;4;.5 & 0.5;1e-4;0.457;16;16 \\
\hline
\textbf{Coffee}  & -3.00;16;0.062 & -3.00;26.337;4096;16 & 8;0.062 & 5.202;512;4 & 0;0.1;16;4 & 0;0.1;300;1024;128 \\
\hline
\textbf{OliveOil}  & -3.00;8;0.5 & -0.82;0.457;256;0.062 & 2;0.125 & 0.457;32;0.125 & 0;0.001;256;32 & 0;0.001;32;32 \\
\hline
\textbf{Beef}  & -3.00;128;0.125 & -3.00;0.685;0.004;16384 & 16;0.016 & 0.457;0.004;16 & 0;1e-4;2;1 & 0;1e-4;0.135;0.004;16 \\
\hline
\end{tabular}
\end{table*}

\begin{table*}[ht]
 \caption{Comparative study using the UCR datasets: classification error rates (in \%) obtained using the first near neighbor classification rule
 and a SVM classifier for the $erp$, ${REDK}_{erp}$, $dtw$ and ${REDK}_{dtw}$ kernels. Two scores are given S1$|$S2: the first one, S1, is evaluated on the training data, while the second one, S2, is evaluated on the test data. Bold faces indicates the lowest error rate between the pairs of  SVMs ($K_{erp}$, $REDK_{erp}$) and ($K_{dtw}$, $REDK_{dtw}$).}
\label{Tab4}
\centering
\begin{tabular}{|l|c|c|c||c|c|c|c|}
\hline
\textbf{DATASET} &  \textbf{1-NN $\delta_{erp}$} & \textbf{SVM $\delta_{erp}$} & \textbf{SVM ${REDK}_{erp}$} & \textbf{1-NN $\delta_{dtw}$} & \textbf{SVM $\delta_{dtw}$} & \textbf{SVM ${REDK}_{dtw}$} \\
\hline\hline
\textbf{Synthetic control} & 0.67$|$3.7 & \textbf{0.33$|$1} & \textbf{0.33}$|$\textbf{1} & 0$|$1.33 & \textbf{0}$|$1.33 & \textbf{0$|$1}  \\
\hline
\textbf{Gun-Point} & 6.12$|$4 & \textbf{0$|$1.33} & \textbf{0$|$1.33} & 18.36$|$9.3 & 6$|$8 & \textbf{0$|$.67} \\
\hline
\textbf{CBF} & 0$|$0.33 & \textbf{3.33}$|$3.44 & \textbf{3.33$|$2.67} & 0$|$0.33 & 3.33$|$1.67 & \textbf{0$|$0.22} \\
\hline
\textbf{Face (all)} & 10.73$|$20.18 & \textbf{.71}$|$17.04 & \textbf{.71}$|$\textbf{16.98} & 6.8$|$19.23 & 4.29$|$14.97 & \textbf{1.25}$|$\textbf{7.89} \\
\hline
\textbf{OSU Leaf} & 30.15$|$40.08 & \textbf{18.5}$|$31.41 & 22$|$\textbf{29.75} & 29$|$44.63 & 28$|$34.3 & \textbf{21$|$25.62} \\
\hline
\textbf{Swedish Leaf} & 11.02$|$12 & 9.2$|$7.68 & \textbf{6.8$|$6.24} & 24.65$|$20.8 & 20.6$|$17.92 & \textbf{6.8$|$6.72} \\
\hline
\textbf{50 Words} & 19.38$|$28.13 & \textbf{16.22$|$16.26} & 16.89$|$17.80 & 33.18$|$31 & 30.89$|$38.90 & \textbf{16.44$|$25.93} \\
\hline
\textbf{Trace} & 10.01$|$17 & \textbf{0$|$0} & \textbf{0}$|$1 & 0$|$0 & \textbf{0$|$0} & \textbf{0$|$0} \\
\hline
\textbf{Two Patterns} & 0$|$0 & \textbf{0$|$0} & \textbf{0$|$0} & 0$|$0 & \textbf{0$|$0} & \textbf{0$|$0} \\
\hline
\textbf{Wafer} & .1$|$0.9 &.2$|$0.62 & \textbf{0.1$|$0.34} & 1.4$|$2.01 & 2.4$|$3.11 & \textbf{0.2$|$0.42} \\
\hline
\textbf{face (four)} & 4.35$|$10.2 & \textbf{4.17$|$3.41} & 8.33$|$4.55 & 26.09$|$17.05 & 12.5$|$25 & \textbf{8.33$|$3.41} \\
\hline
\textbf{Ligthing2} & 11.86$|$14.75 & 13.33$|$26.23 & \textbf{11.67}$|$\textbf{22.95} & 13.56$|$13.1 & 16.67$|$\textbf{8.2} & \textbf{8.33}$|$16.39 \\
\hline
\textbf{Ligthing7} & 23.19$|$30.1 & \textbf{22.86$|$17.81} & \textbf{22.86$|$17.81} & 33.33$|$27.4 & 30.00$|$26.03 & \textbf{12.86$|$17.81} \\
\hline
\textbf{ECG} & 10.01$|$13 & \textbf{7}$|$16 & 9$|$\textbf{12} &  23.23$|$23 & 12$|$19 & \textbf{6$|$11} \\
\hline
\textbf{Adiac} & 35.99$|$37.85 & 33.33$|$31.71 & \textbf{24.10$|$24.4} & 40.62$|$39.64 & 36.41$|$35.29 & \textbf{21.79$|$20.97} \\
\hline
\textbf{Yoga} & 14.05$|$14.7 & 21$|$12.63 & \textbf{11.67$|$10.5} & 16.37$|$16.4 & 20.33$|$17.03 & \textbf{11.67$|$11.47} \\
\hline
\textbf{Fish} & 16.09$|$12 & \textbf{9.14}$|$6.86 & \textbf{9.14$|$4.57} & 26.44$|$16.57 & 25.71$|$22.29 & \textbf{7.43$|$4.57} \\
\hline
\textbf{Coffee} & 25.93$|$25 & \textbf{14.29$|$17.86} & \textbf{14.29$|$17.86} & 14.81$|$17.86 & \textbf{7.14}$|$\textbf{7.14} & \textbf{7.14$|$7.14} \\
\hline
\textbf{OliveOil} & 17.24$|$16.67 & \textbf{13.33$|$16.67} & \textbf{13.33$|$16.67} & 13.79$|$13.33 & \textbf{10$|$13.33} & 13.33$|$\textbf{13.33} \\
\hline
\textbf{Beef} & 68.97$|$50 & \textbf{36.67$|$46.67} & 46.67$|$\textbf{46.67} & 46.67 $|$50 & 33.33$|$53.33 & \textbf{30$|$46.67} \\
\hline
\hline
\textbf{\# Best Scores} & - & \textbf{15}$|$10 & \textbf{15$|$17} & - & 5$|$5 & \textbf{19$|$19} \\
\hline
\textbf{\# Uniquely Best Scores} & - & 3$|$4 & \textbf{5$|$10}& - & 1$|$1 & \textbf{15$|$15} \\
\hline
\end{tabular}
\end{table*}

\begin{figure}[]
\centering
\includegraphics[scale=0.45]{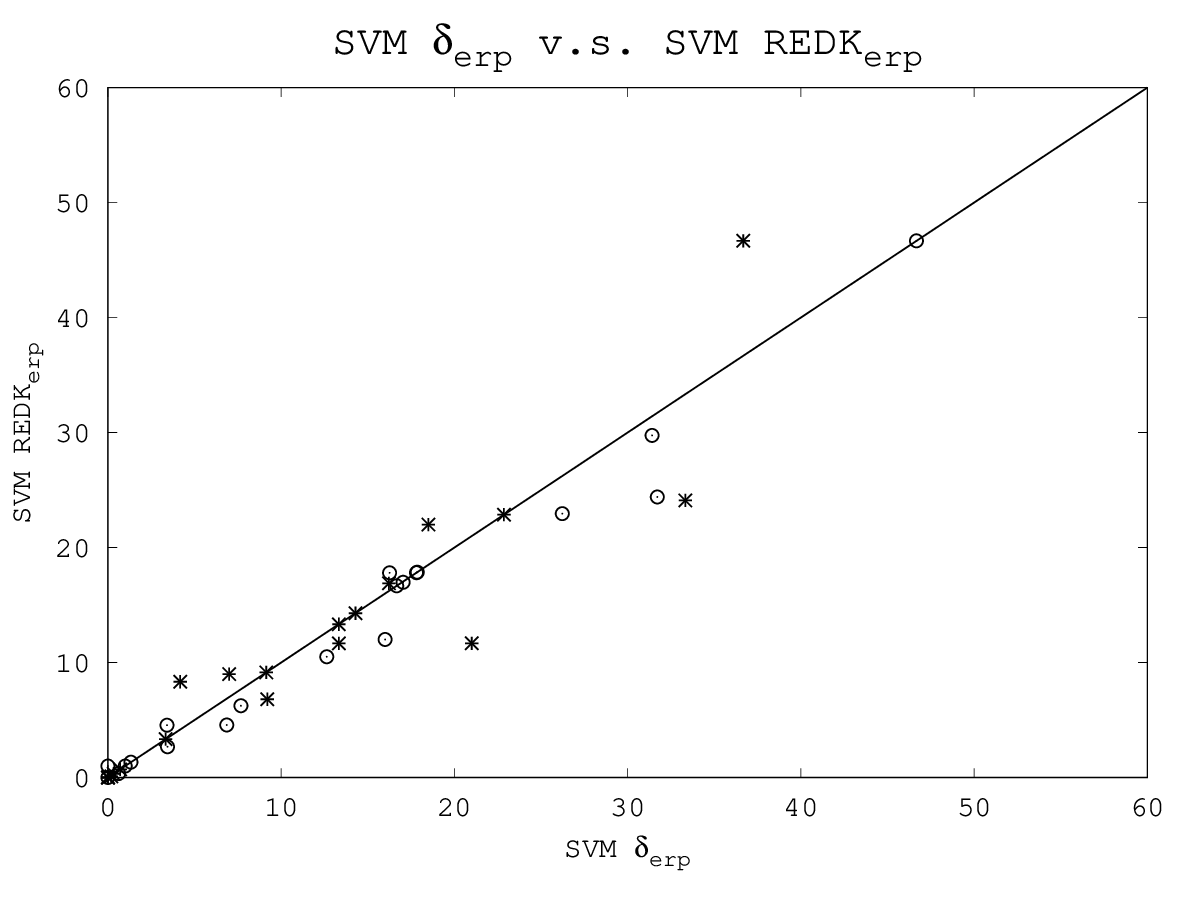}
\caption{Comparison of error rates (in \%) between two SVM classifiers, the first one based on the $\delta_{erp}$ substituting kernel (SVM $\delta_{erp}$), and the second one based on a REDK kernel induced by the $\delta_{erp}$ distance (SVM $REDK_{erp}$). The straight line has a slope of 1.0 and dots correspond, for the pair of classifiers, to the error rates on the train (star) or test (circle) data sets. A dot below (resp. above) the straight line indicates that SVM $REDK_{erp}$ has a lower (resp. higher) error rate than distance SVM $\delta_{erp}$}
\label{fig:SVM_ERP_REDK-ERP}
\end{figure}

\begin{figure}[]
\centering
\includegraphics[scale=0.45]{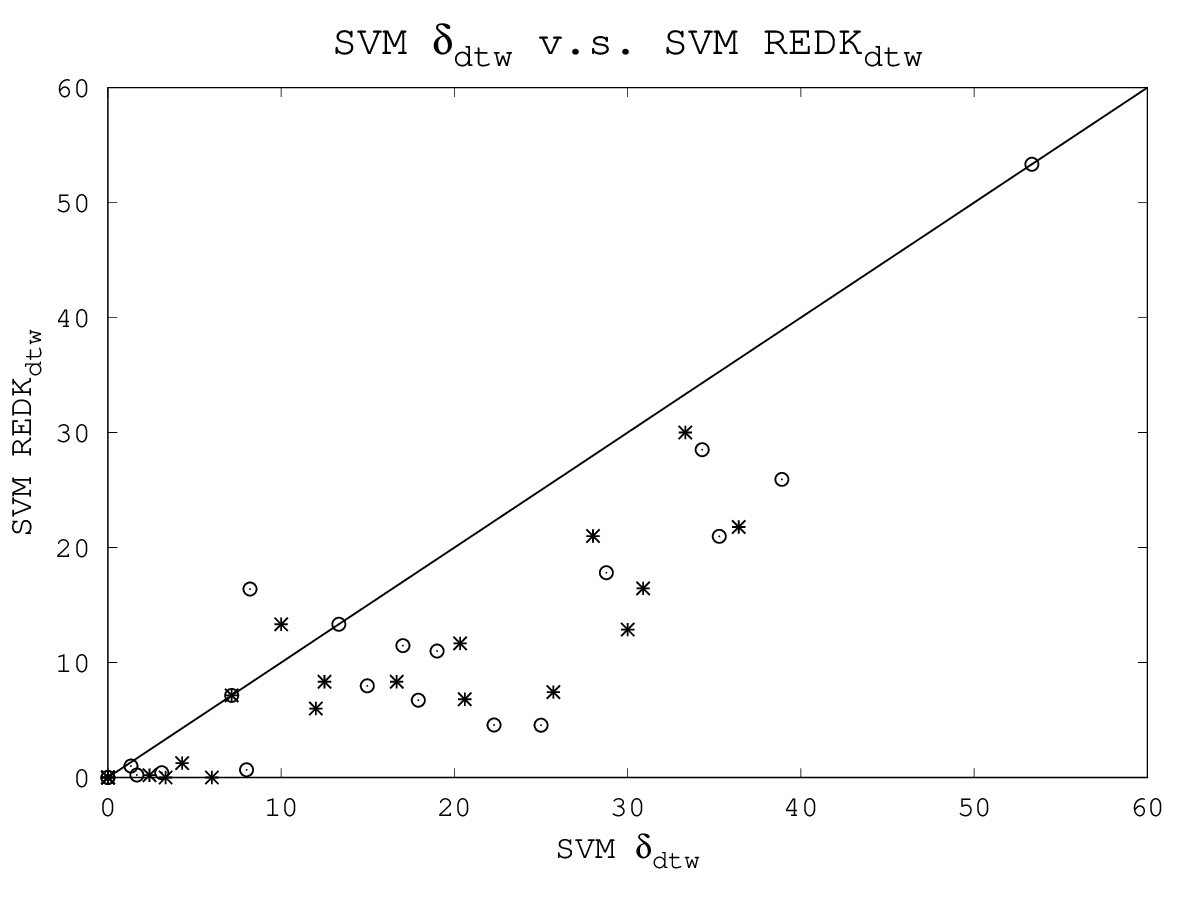}
\caption{Comparison of error rates (in \%) between two SVM classifiers, the first one based on the $\delta_{dtw}$ substituting kernel (SVM $\delta_{dtw}$), and the second one based on a REDK induced by the $\delta_{dtw}$ distance (SVM $REDK_{dtw}$)). The straight line has a slope of 1.0 and dots correspond, for the pair of classifiers, to the error rates on the train (star) or test (circle) data sets. A dot below (resp. above) the straight line indicates that SVM $REDK_{dtw}$ has a lower (resp. higher) error rate than distance $\delta_{dtw}$}
\label{fig:SVM_REDK-DTW}
\end{figure}

\begin{figure}[]
\centering
\includegraphics[scale=0.45]{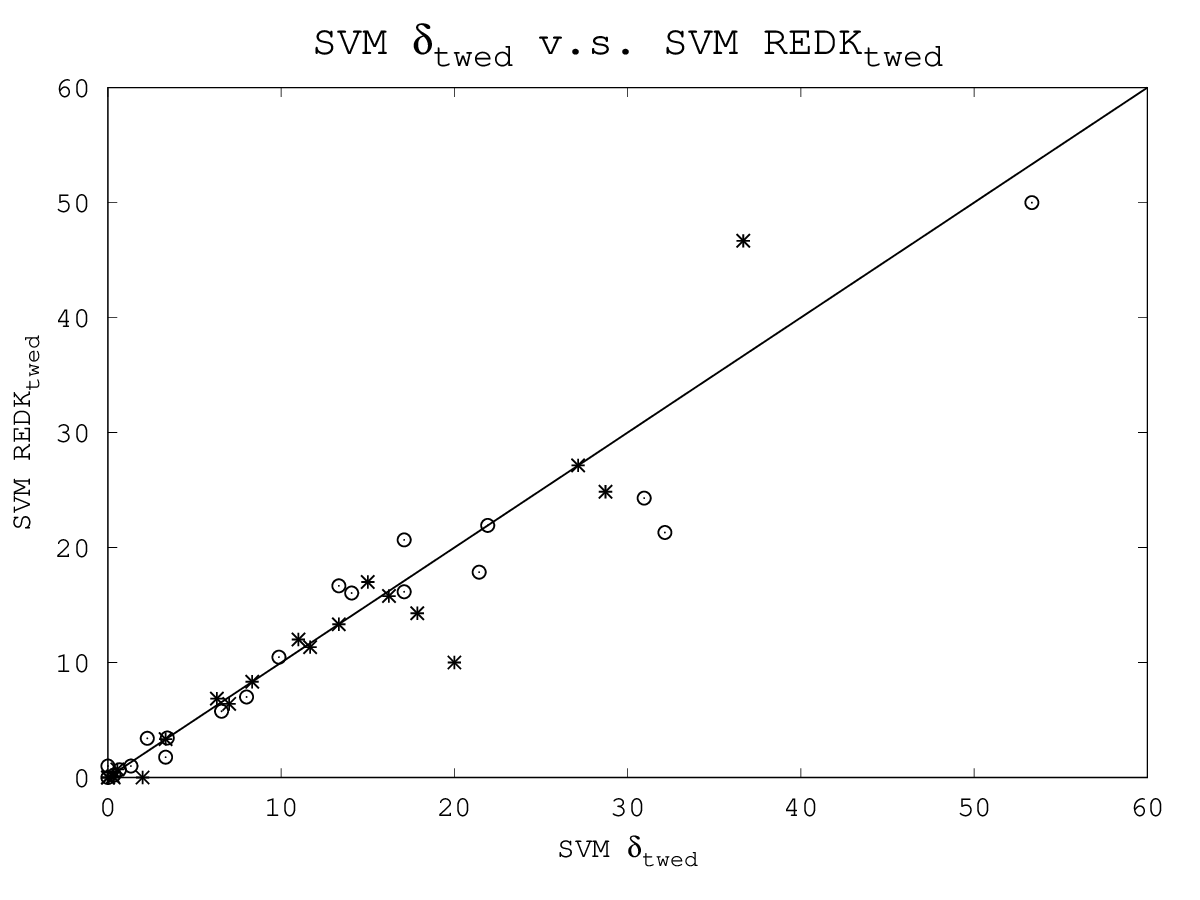}
\caption{Comparison of error rates (in \%) between two SVM classifiers, the first one based on the $\delta_{twed}$ substituting kernel (SVM $\delta_{twed}$), and the second one based on a REDK induced by the $\delta_{erp}$ distance (SVM $REDK_{twed}$). The straight line has a slope of 1.0 and dots correspond, for the pair of classifiers, to the error rates on the train (star) or test (circle) data sets. A dot below (resp. above) the straight line indicates that SVM $REDK_{twed}$ has a lower (resp. higher) error rate than distance SVM $\delta_{twed}$}
\label{fig:SVM_REDK-TWED}
\end{figure}

\subsection{Discussion}

\begin{table*}[t]
 \caption{Comparative study using the UCR datasets: classification error rates (in \%) obtained using the first near neighbor classification rule
 and a SVM classifier for the $\delta_{twed}$ and ${REDK}_{twed}$ kernels. Two scores are given S1$|$S2: the first one, S1, is evaluated on the training data, while the second one, S2, is evaluated on the test data. Bold faces indicates the lowest error rate between the pair of SVMs ($K_{twed}$, $REDK_{twed}$).}
\label{Tab5}
\centering
\begin{tabular}{|l|c|c|c|c|c|c|c|}
\hline
\textbf{DATASET} & \textbf{1-NN $\delta_{twed}$} & \textbf{SVM $\delta_{twed}$} & \textbf{SVM ${REDK}_{twed}$} \\
\hline\hline
\textbf{Synthetic control} & 1$|$2.33 & \textbf{0}$|$1.33 & \textbf{0$|$1} \\
\hline
\textbf{Gun-Point} & 0$|$1.33 & 2$|$\textbf{0.67} & \textbf{0$|$0.67}  \\
\hline
\textbf{CBF} &0$|$0.67 & \textbf{3.33}$|$3.33 &  \textbf{3.33$|$1.77} \\
\hline
\textbf{Face (all)} & 1.43$|$18.93 & \textbf{0.54}$|$17.10 & 0.71$|$\textbf{16.15} \\
\hline
\textbf{OSU Leaf} & 17.59$|$24.79 & \textbf{15$|$17.36} & 17$|$20.66 \\
\hline
\textbf{Swedish Leaf} & 8.82$|$10.24 & 7$|$6.56 & \textbf{6.4$|$5.76} \\
\hline
\textbf{50 Words} & 18.26$|$18.9 & 16.22$|$\textbf{14.07} & \textbf{15.78}$|$16.04 \\
\hline
\textbf{Trace} & 1$|$5 & \textbf{0$|$0} & \textbf{0}$|$1 \\
\hline
\textbf{Two Patterns} & 0$|$0.12 & \textbf{0$|$0} & \textbf{0$|$0} \\
\hline
\textbf{Wafer} & 0.1$|$.86 & 0.3$|$0.34 & \textbf{0.1$|$0.2} \\
\hline
\textbf{face (four)} & 8.7$|$3.41 & \textbf{8.33$|$2.27} & \textbf{8.33}$|$3.41 \\
\hline
\textbf{Ligthing2} & 13.56$|$21.31 & 20$|$32.15 & \textbf{10$|$21.31} \\
\hline
\textbf{Ligthing7} & 24.64$|$24.66 & \textbf{27.14$|$21.92} &\textbf{27.14$|$21.92} \\
\hline
\textbf{ECG} & 13.13$|$10 & \textbf{11}$|$8 & 12$|$\textbf{7} \\
\hline
\textbf{Adiac} & 36.25$|$37.6 & 28.72$|$30.95 & \textbf{24.85$|$24.30} \\
\hline
\textbf{Yoga} & 19.06$|$12.97 &11.67$|$\textbf{9.87} & \textbf{11.33}$|$10.46 \\
\hline
\textbf{Fish} & 12.07$|$5.14 & \textbf{6.29$|$3.43} &6.86$|$\textbf{3.43} \\
\hline
\textbf{Coffee} & 18.52$|$21.43 & 17.86$|$21.43 & \textbf{14.29$|$17.86} \\
\hline
\textbf{OliveOil} & 11.11$|$16.67 &  \textbf{13.33$|$13.33} & \textbf{13.33}$|$16.67 \\
\hline
\textbf{Beef} & 58.62$|$53.3 &\textbf{36.67}$|$53.33 & 46.67$|$\textbf{50} \\
\hline
\hline
\textbf{\# Best Scores} & - & 12$|$10 & \textbf{15$|$14}\\
\hline
\textbf{\# Uniquely Best Scores} & - &  5$|$5 & \textbf{8$|$10}\\
\hline
\end{tabular}
\end{table*}

\begin{center}
\begin{table*}[t]
 \caption{Analysis of the deviation to conditionally definiteness for the distance-matrices associated to the $\delta_{dtw}$, $\delta_{erp}$ and $\delta_{twed}$ distances. We report for each dataset the number of positive eigenvalues ($\#Pev$) relatively to the total number of eigenvalues ($\#Ev$)and the deviation to definiteness estimated as $\Delta_p$ that expresses in \%. The expectation is a single positive eigenvalue, $\#Pev=1$, corresponding to $\Delta_p=0\%$. }
\label{Tab:deviation2definiteness}
\centering
\begin{tabular}{|l|c|c|c|c|c|c|}
\hline
\textbf{DATASET} & \multicolumn{2}{c|}{\textbf{$\delta_{dtw}$}} & \multicolumn{2}{c|}{\textbf{$\delta_{erp}$}} & \multicolumn{2}{c|}{\textbf{$\delta_{twed}$}} \\
\hline
 - & \textbf{\#Pev/\#Ev} & \textbf{$\Delta_p$} & \textbf{\#Pev/\#Ev} & \textbf{$\Delta_p$} & \textbf{\#Pev/\#Ev} & \textbf{$\Delta_p$}  \\
\hline\hline
\textbf{Synthetic control} & 110/300 & 15.66\% & 8/300 & .16\% & 6/300 & .22 \%\\
\hline
\textbf{Gun-Point} & 23/50 & 2.54\% & 1/50& 0\% & 1/50 & 0\% \\
\hline
\textbf{CBF} & 5/30 & 3.36\% &  1/30 & 0\% & 1/30 & 0\% \\
\hline
\textbf{Face (all)} & 242/560 & 26.6\% & 83/560 & 2.42\% & 41/560 & 1.89\%\\
\hline
\textbf{OSU Leaf} & 96/200 & 31.79\% & 29/200 &2.97\% & 16/200& .89\%\\
\hline
\textbf{Swedish Leaf} & 206/500 & 17.04\% &24/500 & .68\%&23/500 & .41\%  \\
\hline
\textbf{50 Words} &218/450 & 34.03\% & 119/450&9.54\% & 93/450& 4.85\% \\
\hline
\textbf{Trace} &43/100 & 5.42\% & 1/100& 0\%& 1/100&0\% \\
\hline
\textbf{Two Patterns} & 453/1000& 36.7\%& 259/1000&13.8\% &226/1000 & 9.85\% \\
\hline
\textbf{Wafer} &497/1000 &14.84\% &137/1000 &1.29\% &39/1000 &.04\% \\
\hline
\textbf{face (four)} &2/24 & .74\%& 1/24&0\% &1/24 & 0\%\\
\hline
\textbf{Ligthing2} & 20/60 & 13.44\% & 1/60& 0\%& 1/60& 0\%  \\
\hline
\textbf{Ligthing7} & 24/70& 14.25\%& 1/70&0\% &1/70 &0\% \\
\hline
\textbf{ECG} &38/100 &14.7\% &1/100 &0\% &1/100 & 0\%\\
\hline
\textbf{Adiac} &159/390 &5.54\% & 26/390 & .82\%& 39/390&.69\% \\
\hline
\textbf{Yoga} & 142/300&23.4\% &29/300 & 3.17\%& 10/300& .41\%\\
\hline
\textbf{Fish} & 71/175& 17.57\%&1/175 &0\% &1/175 &0\% \\
\hline
\textbf{Coffee} & 12/28&8.83\% &1/28 &0\% &1/28 &0\% \\
\hline
\textbf{OliveOil} &4/30 &.24\% &1/30 &0\% &1/30 &0\% \\
\hline
\textbf{Beef} &15/30 &6.17\% &1/30 &0\% &1/30 &0\% \\
\hline
\end{tabular}
\end{table*}
\end{center}


\subsubsection{REDK experiment analysis}

Tables \ref{Tab4} and \ref{Tab5} show the classification error rates obtained for the tested methods, e.g. the first near neighbor classifier based on the $\delta_{erp}$, $\delta_{dtw}$ and $\delta_{twed}$ distances (1-NN $\delta_{erp}$, 1-NN $\delta_{dtw}$ and 1-NN $\delta_{twed}$), the Gaussian RBF kernel SVM based on the same distances (SVM $\delta_{erp}$, SVM $\delta_{dtw}$ and SVM $\delta_{twed}$) and Euclidean distance, and the Gaussian RBF kernel SVM based on the REDK kernels (SVM $REDK_{erp}$, SVM $REDK_{dtw}$ and SVM $REDK_{twed}$). 

In this experiment, we show that the SVM classifiers clearly outperform in general the 1-NN classifiers that are used as a baseline. But the interesting results reported in tables \ref{Tab4} and \ref{Tab5} and figures \ref{fig:SVM_ERP_REDK-ERP}, \ref{fig:SVM_REDK-DTW} and \ref{fig:SVM_REDK-TWED}  is that SVM $REDK_{erp}$ and SVM $REDK_{twed}$ perform slightly better than SVM $\delta_{erp}$ and SVM $\delta_{twed}$ respectively, and the SVM $REDK_{dtw}$ is clearly much efficient than the SVM $\delta_{dtw}$.  This could come from the fact that $\delta_{erp}$ and $\delta_{twed}$ are metrics but not $\delta_{dtw}$, but another explanation could be related to the the SMO optimization involved into the SVM learning.  SVM $\delta_{dtw}$ poorly behaves compared to the other tested classifiers, probably because the SVM optimization process does not perform well. Nevertheless, the $REDK_{dtw}$ kernel based on $\delta_{dtw}$ greatly improves the classification results. 
To further explore the potential impact of indefiniteness on classification rates, we give in table \ref{Tab:deviation2definiteness} two quantified hints of deviation to conditionally definiteness for the distance-matrices corresponding to the $\delta_{dtw}$, $\delta_{erp}$ and $\delta_{twed}$ distances. Since to be conditionally definite (negative) a pairwise distance-matrix should have a single positive eigenvalue, the first hint is the number of positive eigenvalues $\#Pev$ (we give also as a reference the total number of eigenvalues, $\#Ev$). The second hint, $\Delta_p=100*\frac{\sum_{ev_i>0}(ev_i)-\text{ArgMax}_{ev_i>0}\{ev_i\}}{\sum_{ev_i>0}ev_i}$, where $ev_i$ is an eigenvalue of the distance matrix, quantifies the weight of the extra positive eigenvalues relatively to the weight of the total number of positive eigenvalues. Therefore, a conditionally definite (negative) matrix should be such that simultaneously $\#Pev=1$ and $\Delta_p=0$.

By examining the distance-matrices corresponding to each training datasets and for each distances  $\delta_{dtw}(A,B)$, $\delta_{erp}(A,B)$ and $\delta_{twed}(A,B)$, we can show that the $\delta_{dtw}$ kernel is much further away from a conditionally definite matrix than the $\delta_{erp}$ and $\delta_{twed}$ kernels. The distance that is closer to conditional definiteness is the $\delta_{twed}$ distance. This is clearly quantifiable by computing the number of positive eigenvalues, as well as their amplitudes, of the pairwise distance matrices. For datasets such as \textit{FaceAll, OSULeaf, SwedishLeaf, 50words, Adiac, Two\_Patterns, Fish} etc., all the three distances lead to indefinite pairwise distance matrices. The REDK regularization brings in general a significant improvement, specifically when the number and amplitudes of the extra positive eigenvalues are high.  Furthermore, for datasets of small sizes (such as \textit{CBF}, \textit{Beef}, \textit{Coffee}, \textit{OliveOil}, etc.), $\delta_{erp}$ and $\delta_{twed}$  produce  conditionally definite matrices where $\delta_{dtw}$ does not. One can see that the regularization brought by REDK is much more effective on $\delta_{dtw}$ on these data set. Nevertheless, some datasets are better classified by SVM that use directly the distance substituting kernel instead of the derived REDK regularized kernel. 

To conclude, our experiment clearly shows that for problems involving distance matrices \textit{far} from definiteness (more than 5-10\% of eigenvalues are positive, see Table 4), the REDK regularized version outperforms the original elastic distance substituting kernel. This is particularly true for $\delta_{dtw}$. For $\delta_{twed}$, or even $\delta_{erp}$ distances, most of the training sets lead to pairwise distance-matrices that are either conditionally definite negative or close to  CND matrices (with very few extra positive eigenvalues). Nevertheless, in most cases, the REDK-$\delta_{erp}$ and REDK-$\delta_{twed}$ outperformed $\delta_{erp}$ and $\delta_{twed}$ respectively, showing some robustness of the REDK regularization. The main drawback of REDK is the extra parameter $\nu'$. This extra parameter $\nu'$ makes the search for an optimal setting on the train data more difficult and requires more learning data to converge. The trade-off between learning and generalization is therefore more complex to optimize.

\section{Conclusion}

Following the work on convolution kernels \cite{TWED:Haussler99} and local alignment kernels defined for string processing around the Smith and Waterman algorithm \cite{TWED:SmithWaterman81} \cite{TWED:Saigo04}, or defined for time series on the basis of the DTW similarity measure \cite{TWED:Cuturi07}, we have addressed the definiteness of elastic distances through the construction of positive definite kernels built from the recursive definitions of elastic (editing) distances themselves and achieve some extension and generalization of these previous results. 

Contrary to other approaches that rely on the regularization of the Gram-matrix \cite{TWED:Wu2005}\cite{Chen:jmlr2009}, this approach does not depend at all on the data, excpet for the optimization of a regularizing parameter ($\nu'$). By adding an extra recursive term we achieve some simple sufficient conditions, that are weaker than those proposed in \cite{TWED:Saigo04} or \cite{TWED:Cuturi07}, allowing to build positive definite exponentiated REDK. These conditions are basically satisfied by any classical elastic distance defined by a recursive equation. In particular they can be applied to the edit distance, the well known Dynamic Time Warping measure and to some variants such as the Edit Distance With Real penalty and the Time Warp Edit Distance, the latter two being metrics as well as the symbolic edit distance. 
Furthermore, our results apply when a symmetric \textit{corridor} is used to limit the search space required to evaluate the elastic distance, thus reducing consequently the computation time, which, in the worse case, i.e. without corridor, remains quadratic.

The experiments conducted on a variety of time series datasets show that the  positive definite REDKs outperforms in general the indefinite elastic distances from which they are derived, when considering 1-NN and SVM classification tasks. This is mostly striking when the Gram-matrix evaluated on the train data sets  is \textit{far} from definiteness, which is the case when DTW is used. Recent experiment in isolated gesture recognition \cite{marteau:2014} corroborate this observation.

\appendices
\section{Indefiniteness of some elastic measures}
\label{Appendix:A}
\subsection{The Levenshtein distance}
The Levenshtein distance kernel $\varphi(x,y)=\delta_{lev}(x,y)$ is known to be indefinite. Below, we discuss the first known counter-example produced by \cite{TWED:Cortes03}.
Let us consider the subset of sequences $V=\{abc, bad, dab, adc, bcd\} $ that leads to the following distance matrix
\begin{equation}
 M_{lev}^{V}= \left( \begin{array}{ccccc}
0 & 3 & 2 & 1 & 2 \\
3 & 0 & 2 & 2 & 1 \\
2 & 2 & 0 & 3 & 3\\
1 & 2 & 3 & 0 & 3\\
2 & 1 & 3 & 3 & 0\end{array} 
\right)
\end{equation}

and consider coefficient vectors $C$ and $D$ in $\mathbb{R}^5$ such that\\
$C=\left[1, 1, -2/3, -2/3, -2/3 \right]$ with $\sum_{i=1}^{5} c_i = 0$ and \\
$D=\left[1/3, 2/3, 1/3, -2/3, -2/3 \right]$ with $\sum_{i=1}^{5} d_i = 0$.

Clearly $C  M_{lev}^{V} C^T= 2/3 >0$ and $D M_{lev}^{V} D^T= -4/3 < 0$, showing that  $M_{lev}^{V}$ has no definiteness. \\

\subsection{The Dynamic Time Warping distance}

The DTW kernel  $\varphi(x,y)=\delta_{dtw}(x,y)$  is also known not to be conditionally definite. The following example demonstrates this known result.
Let us consider the subset of sequences $V=\{01, 012, 0122, 01222\} $. 

Then the DTW distance matrix evaluated on $V$ is
\begin{equation}
 M_{dtw}^{V}= \left( \begin{array}{cccc}
0 & 1 & 2 & 3 \\
1 & 0 & 0 & 0 \\
2 & 0 & 0 & 0 \\
3 & 0 & 0 & 0  \end{array} 
\right)
\end{equation}
and consider coefficient vectors $C$ and $D$ in $\mathbb{R}^4$ such that\\
$C=\left[1/4, -3/8, -1/8, 1/4 \right]$ with $\sum_{i=1}^{4} c_i = 0$ and \\
$D=\left[-1/4, -1/4, 1/4, 1/4 \right]$ with $\sum_{i=1}^{4} d_i = 0$. 
Clearly $C M_{dtw}^{V} C^T= 2/32 >0$ and $D M_{dtw}^{V} D^T= -1/2 <0$, showing that  $M_{dtw}^{V}$ has no definiteness. \\

\subsection{The Time Warp Edit Distance}
Similarly, it is easy to find simple counter examples that show that TWED kernels are not definite. 

Let us consider the subset of sequences $V=\{010, 012, 103, 301, 032, 123, 023, 003, 302, 321\} $.

For the TWED metric, with $\nu=1.0$ and $\Omega=0.0$ we get the following matrix:
\begin{eqnarray}
 M_{twed}^{V}= & \nonumber \\ 
\left( \begin{array}{cccccccccc}
0 &2 &7 &9  &6  &7  &5  &5  &10 &9 \\
2 &0 &5 &9  &4  &5  &3  &3  &8  &9 \\
7 &5 &0 &6  &7  &4  &6  &2  &5  &10 \\
9 &9 &6 &0  &13 &10 &12 &8  &1  &4 \\
6 &4 &7 &13 &0  &5  &3  &5  &12 &9 \\
7 &5 &4 &10 &5  &0  &2  &6  &9  &6 \\
5 &3 &6 &12 &3  &2  &0  &4  &11 &8 \\
5 &3 &2 &8  &5  &6  &4  &0  &7  &10 \\
10 &8 &5 &1 &12 &9  &11 &7  &0  &5 \\
9 &9 &10 &4 &9  &6  &8  &10 &5  &0\\
\end{array} 
\right)
\end{eqnarray}

The eigenvalue spectrum for this matrix is the following:\\
 $\{$ $4.62$, $0.04$, $-2.14$, $-0.98$, $-0.72$, $-0.37$, $-0.19$, $-0.17$, $-0.06$, $-0.03$ $\}$. 
This spectrum contains 2 strictly positive eigenvalues, showing that $M_{twed}^{V}$ has no definiteness. \\

\subsection{The Edit Distance with Real Penalty}
For the ERP metric, with $g=0.0$ we get the following matrix:
\begin{equation}
 M_{erp}^{V}= \left( \begin{array}{cccccccccc}
0 &2 &3 &3 &4 &5 &4 &2 &4 &5 \\
2 &0 &3 &5 &2 &3 &2 &2 &4 &5 \\
3 &3 &0 &4 &3 &2 &3 &1 &3 &4 \\
3 &5 &4 &0 &7 &6 &7 &5 &1 &2 \\
4 &2 &3 &7 &0 &3 &2 &2 &6 &5 \\
5 &3 &2 &6 &3 &0 &1 &3 &5 &4 \\
4 &2 &3 &7 &2 &1 &0 &2 &6 &5 \\
2 &2 &1 &5 &2 &3 &2 &0 &4 &5 \\
4 &4 &3 &1 &6 &5 &6 &4 &0 &1 \\
5 &5 &4 &2 &5 &4 &5 &5 &1 &0 \\
\end{array} 
\right)
\end{equation}

The eigenvalue spectrum for this matrix is the following:\\
  $\{$ $4.63$, $0.02$, $1.39e-17$, $-2.21$, $-0.97$, $-0.56$, $-0.41$, $-0.26$, $-0.17$, $-0.08$ $\}$. 
This spectrum contains 3 strictly positive eigenvalues (although the third positive eigenvalue which is very small could be the result of the imprecision of the used diagonalization algorithm), showing that $M_{erp}^{V}$ has no definiteness. \\

\section{Proof of our main result}
\label{Appendix:B}
\subsection{Proof of theorem \ref{theorem:Definiteness of REDK}}

\begin{definition}
\label{def:Un}
Let $\mathbb{U}_n$ be the subset of $\mathbb{U}$ containing all the sequences whose lengths are lower or equal to $n$.\\
\end{definition}

\begin{definition}
\label{def:localKernels}
For all $(a,b) \in ((S \times T) \cup \{\Omega\})^{2}$ let $\kappa(a,b)$ be the local kernel defined as follows:\\
$\kappa(a,b)=f(\Gamma(a \rightarrow b))$\\
where $\Omega$ stands for the \textit{null} sequence.
\end{definition}

\begin{definition}
\label{def:piMap}
Let $\pi$ be an ordered alignment map between two finite non empty sequences of successive integers of the form ${0,..,n}$. Basically $\pi$ is a finite sequence of pairs of integers $\pi(l)=(i_l, j_l)$ for $l \in \{0,..., |\pi|-1\}$, satisfying the following conditions
\begin{itemize}
 \item $0 \le i_l, \forall l \in {0, .., |\pi|-1}$
 \item $i_{l} \le i_{l-1}+1, \forall l \in {1, .., |\pi|-1}$ 
 \item $j_{l} \le j_{l-1}+1, \forall l \in {1, .., |\pi|-1}$
 \item $i_{l-1} < i_l$ or $j_{l-1} < j_l, \forall l \in \{1, .., |\pi|-1\}$
\end{itemize}
$\pi_x(l)=i_l$ and $\pi_y(l)=j_l$ are the two coordinate access functions for the $l^{th}$ pair of mapped integers so that $\pi(l)=(\pi_x(l), \pi_y(l))$. 
 
For all $n \ge 1$, let $\mathcal{M}_n$ be the set of alignment maps $\pi$ such that the two sets of integers mapped by $\pi$ are $\{1 \cdots n\} \times \{1 \cdots n\}$. By convention we set $\mathcal{M}_0=\emptyset$.\\

\end{definition}

\begin{figure}[]
\centering
\includegraphics[scale=0.4]{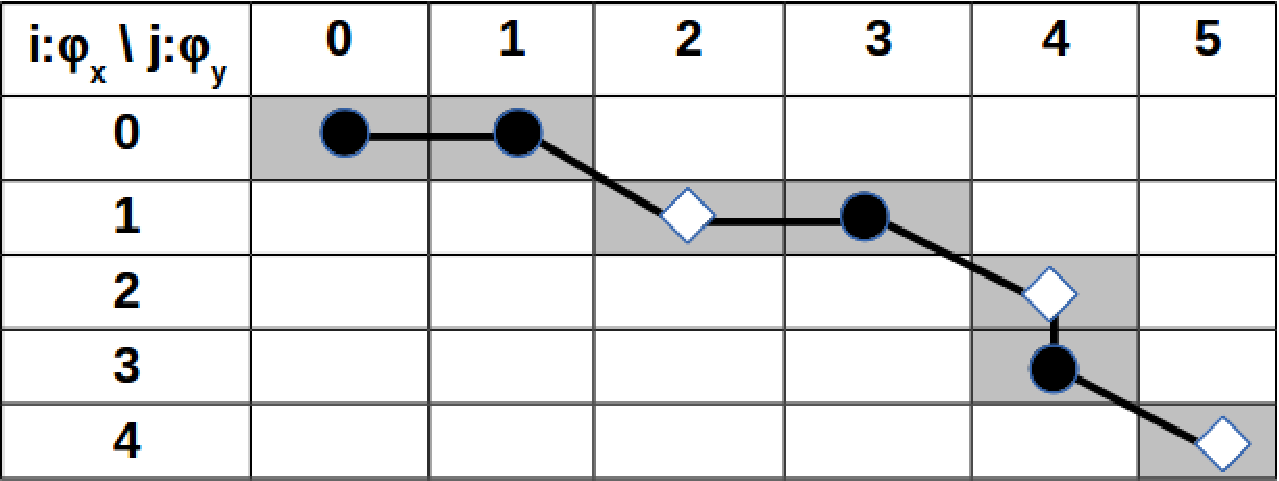}
\caption{Example of an alignment path corresponding to the alignment map $(0,0)(0,1)(1,2)(1,3)(2,4)(3,4)(4,5)$. The white squares correspond to substitution operations and black circles to either deletion or insertion operations.}
\label{fig:piMathPath}
\end{figure}

As shown in figure \ref{fig:piMathPath}, there exists a direct correspondence between an alignment map and an alignment path, i.e. a finite sequence of editing operations.\\

\begin{definition}
\label{piEmbeddings}
For all $n$ we introduce two projections, basically two vectorized representations,  $\varphi_{\pi_x}: \mathbb{U}_n \rightarrow ((S \times T) \cup \{\Omega\})^{2n}$ and $\varphi_{\pi_y}: \mathbb{U}_n \rightarrow ((S \times T) \cup \{\Omega\})^{2n}$ induced uniquely by any alignment map $\pi \in \mathcal{M}_n$. \\ 

The principle for the construction of these two unique projections, given any sequence $A$ and any alignment map $\pi \in \mathcal{M}_n$, is straightforward. With the convention that if $k>|A|$ then $A(k)=\Omega_A(|A|)$ and $\Omega_A(k)=\Omega_A(|A|)$, during the course along $\pi$ of length $L=|\pi|$ at step $l$, $l \in \{1 \cdots L\}$, we apply the following rules:
\begin{enumerate}[i)]
\item if both indexes $\pi_x(l)$ and $\pi_y(l)$ increase 
, then, we insert $A(\pi_x(l))$ in $\varphi_{\pi_x}(A)$  and  $A(\pi_y(l))$ in $\varphi_{\pi_y}(A)$.\\
\item if only index $\pi_x(l)$ increases, then we insert $A(\pi_x(l))$ in $\varphi_{\pi_x}(A)$ and $\Omega_A(\pi_y(l))$ in $\varphi_{\pi_y}(A)$, with the convention that, if $\pi_x(l)>|A|$, $A(\pi_x(l))=\Omega_A(\pi_x(l))$  \\
\item if only index $\pi_y(l)$ increases, then we insert $\Omega_A(\pi_x(l))$ in $\varphi_{\pi_x}(A)$ and $A(\pi_y(l))$ in $\varphi_{\pi_y}(A)$. \\
\item when we reach the end of $\pi$, if the lengths  of $\varphi_{\pi_x}(A)$ (respectively  $\varphi_{\pi_y}(A)$) is shorter than $2n$, then we insert $\Omega$ into the remaining dimensions.\\
\end{enumerate}

For example, for a sequence $A$ of length $5$ the two projections in $((S \times T) \cup \{\Omega\})^{2 \times 5}$ corresponding to the alignment path depicted in Figure \ref{fig:piMathPath} are\\
$\varphi_{\pi_x}(A) = \Omega(1) \textbf{A(1)}  \Omega(2)  \textbf{A(2)}  A(3)  \textbf{A(4)} \Omega \Omega \Omega \Omega$ \nonumber\\ 
$\varphi_{\pi_y}(A) = A(1)      \textbf{A(2)}  A(3)       \textbf{A(4)} \Omega(5) \textbf{A(5)} \Omega \Omega \Omega \Omega$ \nonumber 

where $\Omega(i)=\Omega_A(i)$ is the symbol used for an insertion or deletion operation. This symbol depends on the elastic distance. For DTW,  $\Omega(i)=A(i)$ if $0 < i \le |A|$, or $\Omega(i)=\Omega$ if $i> |A|$.

Then, for any $A \in \mathbb{U}_n$ and $\pi \in \mathcal{M}_n$, we call $\mathcal{P}_{\pi}(A)=\{\varphi_{\pi_x}(A),\varphi_{\pi_y}(A)\}$ the set of projections (or parts) for sequence $A$ induced by $\pi$. Note that all these projections are sequences whose lengths are $2n$. Even 
%
%
\end{definition}

\begin{proposition}
\label{prop:globalPathKernel}
If the local kernel $k(x,y)=f(\Gamma(x \rightarrow y))$ is positive definite on $\left((S \times T) \cup \{\Omega\}\right)^2$ then $\forall n \ge 1$ and $\forall \pi \in \mathcal{M}_n$
\begin{equation}
  K_{\pi}(A,B)=\sum_{\varphi(A) \in \mathcal{P}_{\pi}(A)} \sum_{\varphi(B) \in \mathcal{P}_{\pi}(B)} \prod_{p=1}^{2n} k(\varphi(A)_p, \varphi(B)_p) \\
\end{equation}
is a p.d. kernel on $(\mathbb{U}_n)^{2}$. \\
\end{proposition}

Proof of lemma \ref{prop:globalPathKernel} is a direct consequence of the Haussler's \textit{R-convolution} kernel theorem \cite{TWED:Haussler99}. Indeed, since $k(x,y)$ is a p.d. kernel on $\left((S \times T) \cup \{\Omega\}\right)^2$, and, considering the sets of parts  $\mathcal{P}_{\pi}(A)$ and  $\mathcal{P}_{\pi}(B)$ associated respectively to the sequences $A$ and $B$,  the conditions for the Haussler's \textit{R-convolution} are satisfied.\\

Note that $K_{\pi}(A,B)$ simply rewrites as
\begin{align}
K_{\pi}(A,B)=\prod_{p=1}^{2n}k(\varphi_{\pi_x}(A)_p, \varphi_{\pi_y}(B)_p))\nonumber \\
+\prod_{p=1}^{2n}k(\varphi_{\pi_y}(A)_p, \varphi_{\pi_x}(B)_p))\nonumber \\
+\prod_{p=1}^{2n}k(\varphi_{\pi_x}(A)_p, \varphi_{\pi_x}(B)_p))\nonumber \\ 
+\prod_{p=1}^{2n}k(\varphi_{\pi_y}(A)_p, \varphi_{\pi_y}(B)_p))\nonumber  \\
\end{align}

Let $\mathcal{M}_{n,h} \subset \mathcal{M}_{n}$ be the subset of paths that are restricted by the symmetric corridor induced by any symmetric indicator function $h$ entering into the construction of the RAfP functions  $\mathcal{C}_{h}(.,.)$ and $\mathcal{C}_{h\delta}(.,.)$.

\begin{proposition}
\label{prop1}
For any $n \in \mathbb{N}$, any $\pi \in \mathcal{M}_{n,h}$, and any $(A,B) \in (\mathbb{U}_n)^2$,  we have 
\begin{equation}
 \mathcal{K}(A,B) = \frac{1}{2}\sum_{\pi \in \mathcal{M}_{n,h}} K_{\pi}(A,B)
\end{equation}
\end{proposition}

We state first that $\forall \pi \in \mathcal{M}_{n,h}$, $\exists! \tilde{\pi} \in \mathcal{M}_{n,h}$ such that $\pi_x = \tilde{\pi}_y$ and $\pi_y = \tilde{\pi}_x$. ($\pi$, $\tilde{\pi}$) is a pair of symmetrical paths with respect to the main diagonal.

Thus, since $\mathcal{C}_{h}(.,.)$ evaluates the sum of the \textit{cost-product} along all paths in $\mathcal{M}_{n,h}$, that is
\begin{align}
\mathcal{C}_{h}(A,B)=\sum_{\pi \in \mathcal{M}_{n,h}} \prod_{p=1}^{2n} k(\varphi_{\pi_x}(A)_p, \varphi_{\pi_y}(B)_p)
\end{align}
we can rewrite
\begin{align}
\mathcal{C}_{h}(A,B)=\sum_{\pi \in \mathcal{M}_{n,h}} \Bigg(\frac{1}{2}\prod_{p=1}^{2n} k(\varphi_{\pi_x}(A)_p, \varphi_{\pi_y}(B)_p) \nonumber\\
+\frac{1}{2}\prod_{p=1}^{2n} k(\varphi_{\tilde{\pi}_x}(A)_p, \varphi_{\tilde{\pi}_y}(B)_p)\Bigg) \nonumber\\
=\frac{1}{2}\sum_{\pi \in \mathcal{M}_{n,h}} \Bigg(\prod_{p=1}^{2n} k(\varphi_{\pi_x}(A)_p, \varphi_{\pi_y}(B)_p) \nonumber\\
+\prod_{p=1}^{2n} k(\varphi_{\pi_y}(A)_p, \varphi_{\pi_x}(B)_p)\Bigg)\nonumber
\end{align}

Similarly, for $\mathcal{C}_{h\delta}(.,.)$ we get
\begin{align}
\mathcal{C}_{h,\delta}(A,B)=\frac{1}{2}\sum_{\pi \in \mathcal{M}_{n,h}} \Bigg(\prod_{p=1}^{2n} k(\varphi_{\pi_x}(A)_p, \varphi_{\pi_x}(B)_p) \nonumber\\
+\prod_{p=1}^{2n} k(\varphi_{\pi_y}(A)_p, \varphi_{\pi_y}(B)_p)\Bigg) \nonumber
\end{align}

Hence, the expected result.\\

\textbf{Proof of Proposition \ref{theorem:Definiteness of REDK} (Definiteness of REDK)}\\

Proposition \ref{prop1} states that $\mathcal{K}(.,.)$ is a finite sum of p.d. kernels defined on $(\mathbb{U}_n)^{2}$. According to the closure properties under the addition of p.d. kernels, we show that $\mathcal{K}(.,.)$ is a p.d. kernel defined on $(\mathbb{U}_n)^{2}$. 

As this result is true for all $n$, one can see $\mathcal{K}(.,.)$ as a point-wise limit as $n$ tends toward infinity of a p.d. kernel defined on $(\mathbb{U}_n)^{2}$. This establishes that $\mathcal{K}(.,.B)$ is a p.d. kernel on $\mathbb{U}^{2}$ $\square$.

\ifCLASSOPTIONcompsoc
  \section*{Acknowledgments}
\else
  \section*{Acknowledgment}
\fi
The authors thank the French Ministry of Research, the Brittany Region, the General Council of Morbihan and the European Regional Development Fund that had partially fund this research. The authors also thank Dr. E. Keogh's team at UC Riverside for making available the time series data sets that have been used in this study.




%
\bibliography{RecursiveEditDistanceKernels}{}
\bibliographystyle{IEEEtran}


%







\end{document}